\documentclass[10pt,journal,cspaper,compsoc]{IEEEtran}

\ifCLASSOPTIONcompsoc
\else
\fi

\usepackage[english]{babel} 
\usepackage{tikz}
\usepackage{graphicx}
\usetikzlibrary{calc,shapes.multipart,chains,arrows,fit,shapes,calc,backgrounds,trees,matrix}
\usepackage{url}
\usepackage[utf8]{inputenc}
\usepackage{epsfig}
\usepackage{calc}
\usepackage{amssymb}
\usepackage{amstext}
\usepackage[cmex10]{amsmath}
\usepackage{amsthm}
\usepackage{hyperref}
\usepackage{multirow}
\usepackage{booktabs}
\usepackage{subcaption}

\usepackage{fixltx2e}

\begin{document}

\title{Why Local Search Excels in Expression Simplification}
\author{Ben~Ruijl, Aske~Plaat, Jos~Vermaseren, Jaap~van~den~Herik
\IEEEcompsocitemizethanks{\IEEEcompsocthanksitem Ben Ruijl and Jos Vermaseren are with Nikhef, Science Park 105, 1098 XG Amsterdam, The Netherlands
\IEEEcompsocthanksitem Ben Ruijl, Aske Plaat and Jaap van den Herik are with Leiden University, Niels Bohrweg 1, 2333 CA Leiden, The Netherlands}
\thanks{}}

\markboth{}
{Ruijl \MakeLowercase{\textit{et al.}}: Stochastic Local Search for Simplifying Large Expressions}

\date{\today}

\IEEEcompsoctitleabstractindextext{
\begin{abstract}
Simplifying expressions is important to make numerical integration of large expressions from High Energy Physics tractable. To this end, Horner's method can be used. Finding suitable Horner schemes is assumed to be hard, due to the lack of local heuristics. Recently, MCTS was reported to be able to find near optimal schemes. However, several parameters had to be fine-tuned manually. In this work, we investigate the state space properties of Horner schemes and find that the domain is relatively flat and contains only a few local minima. As a result, the Horner space is appropriate to be explored by Stochastic Local Search (SLS), which has only two parameters: the number of iterations (computation time) and the neighborhood structure. We found a suitable neighborhood structure, leaving only the allowed computation time as a parameter. We performed a range of experiments. The results obtained by SLS are similar or better than those obtained by MCTS. Furthermore, we show that SLS obtains the good results at least 10 times faster. Using SLS, we can speed up numerical integration of many real-world large expressions by at least a factor of 24. For High Energy Physics this means that numerical integrations that took weeks can now be done in hours.
\end{abstract}

}

\maketitle


\section{Introduction}
\IEEEPARstart{S}{implifying} large expressions is important to make numerical integration tractable. In high energy physics, expressions with millions of terms arise from the calculation of Feynman diagrams. The results are used to compare predictions of Quantum Field Theory to the outcomes of particle collision experiments, for example at CERN or Fermilab. Numerical integration for current-generation processes and for next-generation processes required for, e.g., the International Linear Collider will take months. Thus it is important to save computation time by simplifying the underlying expressions.

The standard method for reducing the number of operations uses Horner's rule of lifting variables outside brackets. This reduces the number of multiplications \cite{Horner1819,Kuipers2013}. Afterwards, common subexpression elimination may be applied to reduce the number of operations even further \cite{dragon}. For a multivariate polynomial, the order in which variables are lifted outside brackets is called a \emph{Horner scheme}. The problem of finding an optimal Horner scheme is NP-hard \cite{Ceberio2004}.

Recent successes with Monte Carlo Tree Search (MCTS) \cite{Coulom2006} have shown that the number of operations of expressions can be reduced by at least a factor of 16 for a set of large, real-world, expressions \cite{Kuipers2013}. In MCTS a tree is built selectively over $N$ iterations, expanding only branches that are deemed worthwhile. However, this method introduces an additional parameter, $C_p$, that governs the amount of exploration versus exploitation and that has to be tuned manually. The suggested algorithm SA-UCT (Simulated Annealing - Upper Confidence Bounds applied to Trees) alleviates the tuning problem, but does not eliminate it \cite{Ruijl2014}. 

In this work we study the state space properties of Horner schemes and will find that the state space is relatively flat and has only a few local minima. This finding will have many consequences. The main result is that a basic algorithm such as Stochastic Local Search (SLS) may be more suited to the Horner scheme problem than more complex algorithms with many tunable parameters. The idea is surprising, since the problem of finding suitable Horner schemes has at least three complications: first, it is NP-hard, second, we do not know of any local heuristics to guide the search, and third, evaluation of a single state is slow; i.e., it takes several seconds for some of our benchmark expressions.

The main contributions of this work are fourfold: (1) it shows a real-world domain where basic stochastic search is more suited than many-parameter search algorithms, (2) it shows that SLS is at least 10 times faster than a manually fine-tuned MCTS, (3) it shows that the Horner scheme state space is flat, and (4) that it usually has only a few local minima.

Our algorithms are implemented in the next release of the open source symbolic manipulation system FORM \cite{Vermaseren2014}.

The paper is structured as follows. Section \ref{sec:background} gives an informative background on simplification methods, section \ref{sec:motivation} provides an overview of related algorithms, and section \ref{sec:setup} gives the experimental setup. Next, section \ref{sec:results} presents the results of SLS versus SA, provides a good neighborhood structure, measures state space properties and compares the performance of SLS to MCTS. Section \ref{sec:conclusion} formulates the conclusion. Section \ref{sec:discussion} contains a discussion and gives a vista on finding other methods of reductions. 

\section{\label{sec:background}Background}
To provide some background, we will now describe two methods to reduce the number of operations, namely Horner schemes and common subexpression elimination \cite{Ruijl2014,Ruijl2014B}, followed by some remark on their interplay and the scaling of the computation time.

\subsection{Horner schemes}
Horner's rule is a classic method to reduce the number of multiplications in a polynomial by lifting variables outside brackets \cite{Horner1819,Knuth1997}. In multivariate polynomials, the order in which the variables are lifted outside brackets is called a \emph{Horner scheme}. For instance:
\begin{equation}
x^2z + x^3y + x^3yz \rightarrow x^2(z + x(y(1 + z)))
\end{equation}
Here, first the variable $x$ is extracted (i.e., $x^2$ and $x$) and second, the variable $y$. The number of multiplications is now reduced from $9$ to $4$. However, the order $x, y$ is chosen arbitrarily. One could also try the order $y, x$:
\begin{equation}
x^2z + x^3y + x^3yz \rightarrow x^2z + y(x^3(1 + z))
\end{equation}
for which the number of multiplications is $6$. Evidently, this is a suboptimal Horner scheme. There are $n!$ orders of extracting variables, where $n$ is the number of variables. The problem of selecting an optimal ordering is NP-hard \cite{Ceberio2004}.

Usually, a heuristic is used that orders the variables according to their frequency in the terms. This heuristic is called ``occurrence order''. However, MCTS, which does not use this heuristic, is able to find Horner schemes that reduce the number of operations by at least 50\% more than the occurrence order on large test expressions (number of operations up to 7\,722\,027) \cite{Kuipers2013}.

\subsection{Common subexpression elimination}
\begin{figure}[h]
\centering
\includegraphics[scale=1]{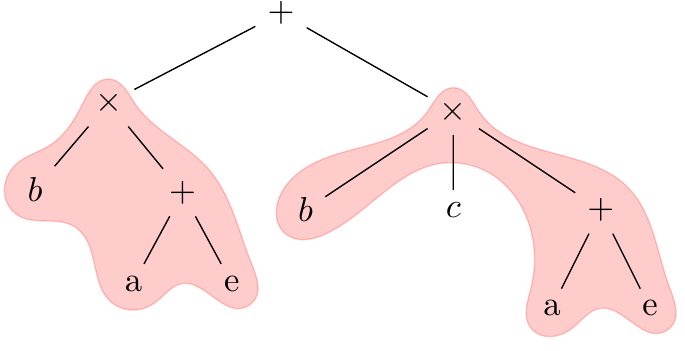}
\caption{A common subexpression (shaded) in an associative and commutative tree representation.}
\label{fig:cse}
\end{figure}

The number of operations can be reduced further by applying common subexpression elimination (CSEE). This method is well known in the field of compiler construction \cite{dragon}, where it is applied to much smaller expressions than in high energy physics, and in the field of computer chess \cite{Adelson1988} where it handles the occurrence of common subtrees by using transposition tables. 
Figure \ref{fig:cse} shows an example of a common subexpression in a tree representation of an expression. The shaded subexpression $b(a+e)$ appears twice, and its removal means removing one superfluous addition and one multiplication.

CSEE is able to reduce both the number of multiplications and the number of additions, whereas Horner schemes are only able to reduce the number of multiplications.

\subsection{Interplay}
We note that there is an interplay between Horner schemes and CSEE: a certain ``optimal'' Horner scheme may reduce the number of multiplications the most, but may expose less common subexpressions than a ``mediocre'' Horner scheme. Thus, we need to find a way to obtain a Horner scheme that reduces the number of operations the most after both Horner and CSEE have been applied.

Finding appropriate Horner schemes is not a trivial task, for at least four reasons. First, there are no known local heuristics. For the Traveling Salesman Problem (TSP), the distance between two cities can be used as a heuristic \cite{Glover1986}, and more specialized heuristics are able to solve symmetric TSP instances with thousands of cities (a historic example is a TSP with 7397 cities \cite{Lin1973,Helsgaun2000}). Second, the Horner scheme is applied to an expression. This means that the scheme has a particular context: the $n$th entry applies to the subexpressions that are created after the first $n-1$ entries in the Horner scheme have been applied to the expression. Third, Horner schemes are asymmetric: a scheme has a well-defined beginning and end. This is, e.g., in contrast with TSPs with closed paths (the most common subclass), since they have circle symmetry (translation and mirror symmetry). Fourth, the evaluation of a Horner scheme and CSEE is slow: for some benchmark expressions the evaluation took multiple seconds on a 2.4~GHz computer (see table \ref{tbl:pol}). Our attempted parallelization of the evaluation function was unsuccessful, since the Horner scheme evaluation function is too fine-grained.

\subsection{Scaling}
The time it takes to apply a Horner scheme is directly related to the number of variables and the number of terms in the expression. The common subexpression elimination time scales linearly with the number of operations. The difficulty of finding a good Horner scheme is related to the size of the permutation space, i.e., related to the number of variables, but also to the distribution of the variables in the terms. The composition of the variables affects the flatness of the state space and the occurrence of saddle points and local minima, as we shall see in section \ref{sec:prop}.

In \cite{Kuipers2013}, and \cite{Ruijl2014}, Monte Carlo Tree Search (MCTS) has been successfully used to find a best candidate of all available Horner schemes. 
In this work we (re)consider which algorithm is best suited for the Horner scheme problem. We discuss six candidate algorithms for the optimization of Horner schemes.

\section{\label{sec:motivation}Related algorithms}
The Horner scheme problem belongs to the class of permutation problems. Many algorithms for optimizing permutation problems have been suggested in the literature. In order to determine which of these algorithms is best suited for Horner schemes, we briefly discuss the characteristics of a selection of six frequently used algorithms: (Stochastic) Local Search \cite{Hoos2004}, Simulated Annealing \cite{Kirkpatrick1983}, Tabu Search \cite{Glover1986}, Ant Colony Optimization \cite{Dorigo1997}, Evolutionary Algorithms \cite{Goldberg1989}, and, Monte Carlo Tree Search \cite{Coulom2006}. Below they are indicated by A to F.

We begin to define the concept of a \emph{neighbor} as it occurs in our problem. The state space of Horner schemes consists of the collection of all possible schemes (a permutation space). If we define transitions between states, then we are able to use graph algorithms to search the state space. All the nodes connected to the current node are called neighbors. There are many options for neighborhood structures (see section \ref{sec:neighb}). In the discussion below we refer to the states reached after swapping two variables in the Horner scheme as a ``neighbor'' (see figure \ref{fig:moves}). Given $n$ variables, there are $n(n-1)/2$ neighbors.



(A) Local Search \cite{Aarts1997,Hoos2004} is a state space exploration method that starts from an initial state and moves to a neighbor of the current state. The task is to find an extreme value. A full local search explores the values of all neighbors and then moves to the neighbor with the best value. For our domain this is impractical, since the number of neighbors is high and a single evaluation takes multiple seconds. A Stochastic Local Search (SLS) \cite{Hoos2004} randomly selects a neighbor and moves if the value of the given neighbor is better than the current value. SLS has two parameters: the number of iterations, $N$, and a neighborhood structure, that defines the transition function. With these characteristics, SLS is a candidate for further research.

(B) Simulated Annealing (SA) \cite{Kirkpatrick1983} is a classic method that is inspired by the removal of crystal defects by the cooling of metals (annealing). It can be viewed as a generalization of Stochastic Local Search, where SA allows transitions to worse states. SA has several parameters, such as the starting temperature, final temperature, and cooling rate. The major difference between SLS and Simulated Annealing is that SA has the ability to escape local minima, whereas SLS, once in such a situation, will remain stuck permanently. Many papers have been published about the tuning of the SA parameters. For instance, in \cite{Miki2003} and \cite{Ben-Ameur2004} suggestions have been made to tune the initial temperature. In section \ref{sec:sls} we measure the performance of SA compared to SLS. 

(C) Tabu Search performs a local search and keeps track of previous visits \cite{Glover1986}. Revisits are (temporarily) disallowed, which allows the search to escape local minima. To improve performance, Tabu Search can be enhanced with short-term, intermediate-term, and long-term memory, at the cost of introducing parameters that have to be tuned manually. In classic Tabu Search, the best neighbor that has not been visited earlier is selected and is added to the tabu list \cite{Glover1986}. However, as mentioned above, evaluating all neighbors is impractical. A Tabu Search that does not try all neighbors introduces problems, since unexplored paths to the global minimum cannot be made anymore through states in the tabu list. Tabu Search may be convenient to escape from deep local minima, but as we will explain in subsection \ref{sec:minima}, we do not encounter these in our search space. Preliminary tests with a basic Tabu Search and only short-term memory did not yield better results. Therefore, we may conclude that the potential benefit (e.g., escaping from deep local minima) does not outweigh the inclusion of additional parameters. Thus, Tabu Search is not a candidate for further research.

(D) Ant Colony Optimization \cite{Dorigo1997} has been successful in optimizing various problems, such as TSP instances. The algorithm has two components: a history component, using information from the performance of previous paths, and a heuristic component that prefers transitions to neighbors that are closest. In TSP the heuristic component could be the distance between two cities. For Horner schemes we do not have local information, since there is no notion of distance between two variables in the scheme. Furthermore, known properties of the underlying expression cannot be used to represent a distance, because the expression itself differs depending on the position in the scheme: the subexpressions to which the variable at position $i$ in the scheme is applied, are constructed using the variables prior in the scheme. Thus, the context of the sub-scheme $x,y$ depends completely on the position in the Horner scheme. Therefore, Ant Colony Optimization is not a candidate for further research.

(E) Evolutionary algorithms work by using mutations and genetic recombinations \cite{Goldberg1989}. The mutations are related to our swap function (see section \ref{sec:neighb}) and the recombinations mean merging parts of two Horner schemes in order to obtain a better one. The recombinations work best if the problem can be split into subproblems. For TSP, it is likely that an optimized subpath is also part of the total shortest path. However, as mentioned above, for Horner schemes the quality of a subpath depends on the context and thus on the variables that have been previously chosen. Therefore, recombinations are not likely to improve the quality of the solutions. Hence, evolutionary algorithms are not a candidate for further research.

(F) Monte Carlo Tree Search (MCTS) \cite{Coulom2006,Browne2012} has been used successfully in finding high-performance Horner schemes \cite{Kuipers2013}. MCTS builds a search tree selectively, where at each level in the tree a new variable is added. Only branches that are deemed profitable are explored further. The number of tree updates $N$ is chosen by the user and is related linearly to the amount of time spent in the search. A commonly used criterion for the selection of the best child, UCT (Upper Confidence bounds applied to Trees), introduces an exploration-exploitation constant $C_p$ that has to be fine-tuned manually \cite{Kocsis2006} (other selection criteria have a similar trade-off parameter). The introduction of Simulated Annealing UCT (SA-UCT), alleviates the tuning of $C_p$, but does not eliminate it \cite{Ruijl2014}. 

Even though MCTS has been successful in finding optimal Horner schemes, we recognize that there are some intrinsic shortcomings to using a tree representation, especially if the depth of the search tree becomes (too) large. We notice that many branches do not reach the bottom when there are more than 20 variables (we recall that the problem depth is equivalent to the number of variables) as is the case with many of our expressions. MCTS determines the scores of a branch by performing a random play-out. If the branch is not constructed all the way to the bottom, the final nodes are therefore random (no optimization). For Horner schemes, the \emph{entire} scheme is important, so sub-optimal selection of variables at the end of the scheme can have a significant impact. In fact, in \cite{Kuipers2013} a parameter is introduced to select whether the Horner schemes should be built forward or backward (the first variables in the scheme are the last to be applied). The backward approach tries to improve results for expressions where the order of the final variables is more sensitive to improvements than the order of the first variables \cite{Ruijl2014}. However, the underlying problem of poor decision making at the end of the tree remains unsolved. Therefore, we do not consider MCTS as a candidate for further research. The issues with tree representations motivated us to look for a method that is symmetric in its optimization: both the beginning and the end have to be optimized equally well.


Using the reasoning above, we may conclude that methods (A) Stochastic Local Search, and (B) Simulated Annealing, are best suited for the Horner scheme problem. Still, we note that the algorithms (C)...(F) mentioned above may be able to outperform SLS and SA, after extensive tuning of their extra parameters. However, we argue that without tuning, these methods do not outperform SLS or SA, since their characteristics are ill-suited to the problem.

So, we continue our investigation with SLS and SA, and we consider the experimental setup for measurements with SLS and SA below.

\section{\label{sec:setup}Experimental setup}
We use eight large benchmark expressions, four from mathematics and four from real-world High Energy Physics (HEP) calculations. In table \ref{tbl:pol} statistics for the expressions are displayed. We show the number of variables, terms, operations, and the evaluation time of applying a Horner scheme and CSEE. 

\begin{table}
\centering
\begin{tabular}{|l|c|c|c|c|}
\cline{2-5}
\multicolumn{1}{c|}{} & variables & terms & operations & eval. time (s)\\
\hline
res(7,4)    & 13 &  2561   &  29\,163 & 0.001\\
res(7,5)    & 14 &  11\,379   &  142\,711 & 0.03\\
res(7,6)    & 15 &  43\,165   &  587\,880 & 0.13\\
res(9,8)	& 19 & 4\,793\,296 & 83\,778\,591 & 25.0\\
\hline
HEP($\sigma$) & 15 & 5716 & 47\,424 & 0.008\\
HEP($F_{13}$) & 24 & 105\,058 & 1\,068\,153 & 0.4\\
HEP($F_{24}$) & 31 & 836\,009 & 7\,722\,027 & 3.0\\
HEP($b$) & 107 & 193\,767 & 1\,817\,520 & 2.0\\
\hline
\end{tabular}
\caption{\label{tbl:pol}The number of variables, terms, operations, and the evaluation time of applying a single Horner scheme and CSEE in seconds, for our eight (unoptimized) benchmark expressions. The time measurement is performed on a 2.4~GHz Xeon computer. All expressions fit in memory (192~GB).}
\end{table}


The expressions called res(7,4), res(7,5), res(7,6), and res(9,8) are resolvents and are defined by $\text{res}(m, n) = \text{res}_x(\sum^m_{i=0} a_ix^i, \sum^n_{i=0} b_ix^i)$, as described in \cite{Leiserson2010}. The number of variables is $m+n+2$. res(9,8) is the largest polynomial we have tested and has been included to test the boundaries of our hardware.


The High Energy Physics expressions represent scattering processes for the International Linear Collider, a likely successor to the Large Hadron Collider. A standard method of calculating the probability of certain collision events is by using perturbation theory. As a result, for each order of perturbations, additional expressions are calculated as corrections to previous orders of precision. The HEP polynomials of table \ref{tbl:pol} are second order corrections to various processes.

HEP($\sigma$) describes parts of the process  $e^+ e^- \rightarrow \mu^+\mu^- \gamma$, namely the collision of an electron and positron that creates a muon, an anti-muon, and a photon.

HEP($F_{13}$), HEP($F_{24}$), and HEP($b$) are obtained from the process $e^+ e^- \rightarrow \mu^+\mu^- u \bar{u}$, namely the collision of an electron and positron that creates a muon, anti-muon, an up-quark, and an up-antiquark. The results can be used to obtain next-generation precision measurements for electron-positron scattering.


\section{\label{sec:results}Results}
In this section we present the results of our measurements on the benchmark polynomials. In section \ref{sec:sls} we measure the difference between Stochastic Local Search and Simulated Annealing.
In section \ref{sec:neighb}, we study the effects of the two parameters of SLS: the number of iterations and the neighborhood structure. 
In section \ref{sec:prop} we investigate two state space properties, namely the occurrence of local minima and the flatness of the state space.
In section \ref{sec:performance} we compare the performance of the Horner schemes that are found by SLS to the results found by MCTS.

\subsection{\label{sec:sls}SLS vs. SA}
A Stochastic Local Search has two parameters: the number of iterations $N$, and the neighborhood structure, which defines the transition function \cite{Aarts1997}. A Stochastic Local Search only moves to a neighbor if the evaluation score (number of operations) is improved. As a consequence, SLS could get stuck in local minima. Therefore, we seriously considered to use Simulated Annealing instead of SLS, since SA has the ability to escape from local minima. 

Simulated Annealing is a popular generalization of SLS. It has four additional parameters, namely the initial temperature $T_i$, the final temperature $T_f$, the acceptance scheme, and the cooling scheme \cite{Kirkpatrick1983}. The temperature governs the probability of accepting transitions with an energy higher than the energy of the current state. The cooling scheme governs how fast and in what way the temperature is decreased during the simulation (linearly, exponentially, etc.). Exponential cooling is frequently used. The acceptance scheme is most often the Boltzmann probability $\exp(\Delta E / T)$, that defines the probability of selecting a transition to an inferior state, given the difference in evaluation score $\Delta E$. 


For SA we consider the following.
If the initial temperature is high, transitions to inferior states are permitted, allowing an escape from local minima. In order to determine the effect of the initial temperature $T_i$ on the results, we will perform a sensitivity analysis. We use Boltzmann probability as the acceptance scheme, exponential cooling, a final temperature of $0.01$, $N=1000$ iterations, and a swap neighborhood structure (for a visualization, see figure \ref{fig:moves}).

\begin{figure}
\centering
\includegraphics[scale=0.4]{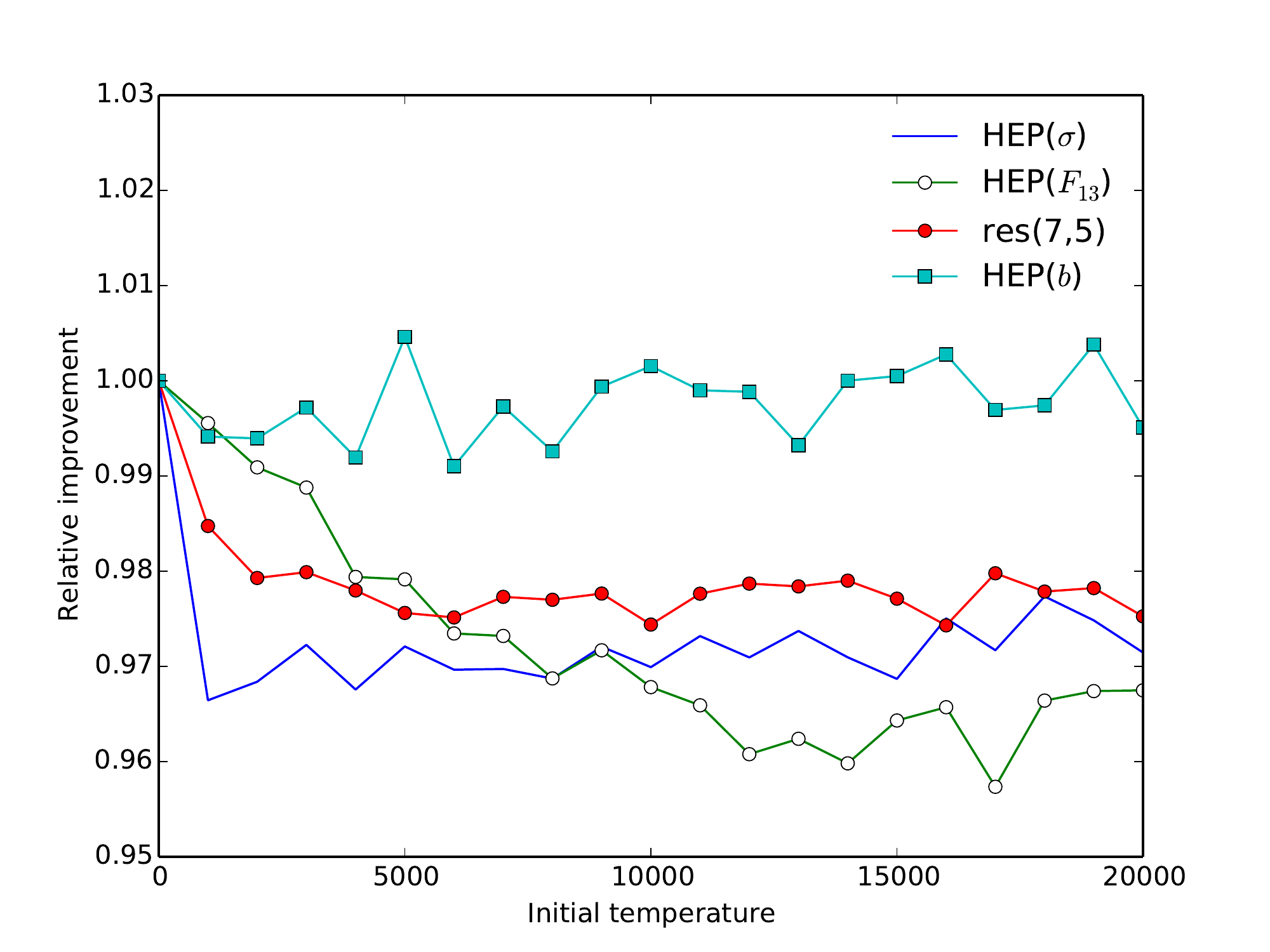}
\caption{The relative improvement (smaller is better) of the number of operations for a given initial temperature $T_i$, compared to $T_i=0$. Each data point is the average of more than 100 SA runs with 1000 iterations, and a swap neighborhood structure. We show the expressions HEP($\sigma$), HEP($F_{13}$), res(7,5), and HEP($b$). The number of operations is only slightly influenced by the initial temperature, since the best improvement over $T_i=0$ is smaller than 5\%.}
\label{fig:tmax}
\end{figure}

In figure \ref{fig:tmax} we show the relative improvement (smaller is better) of the number of iterations for a given initial temperature $T_i$ compared to $T_i=0$ for the expressions HEP($\sigma$), HEP($F_{13}$), res(7,5), and HEP($b$). Naturally, for $T_i=0$, the relative improvement to itself is $1$. For all expressions except HEP($b$), we see a region where the improvement is largest: for HEP($\sigma$) it is approximately $[1000,7000]$, for HEP($F_{13}$) it is $[12\,000,17\,000]$ and for res(7,5) it is $[5000,20\,000]$. This improvement is less than 5\%. For higher $T$, too many transitions to inferior states are accepted to obtain good results. HEP($b$) seems to be independent of the initial temperature. The fluctuations of 1\% are statistical fluctuations. 

The difference between the best results for all the expressions in figure \ref{fig:tmax} and the result at $T=0$ is less than 5\%. Since a $T_i=0$ SA search is effectively an SLS search, this means that an almost parameterless Stochastic Local Search (SLS) is able to obtain results that are only slightly inferior. This is surprising, since the way SLS traverses the state space is different from the way by SA. We here reiterate once more, SLS can get stuck in local minima, whereas SA has the possibility to escape. Furthermore, if a saddle point is reached, SA is able to climb over the hill, whereas SLS has to walk around the hill in order to escape. In subsection \ref{sec:minima} we will show that local minima are sparse, and that most of them are actually saddle points (i.e., local ``minima'' with a way to escape). Consequently, SA performs slightly better not because it can escape from local minima, but because, for some polynomials, walking over a saddle point (SA) is slightly faster to find better states than trying to circumvent the saddle point (SLS). 


The reason why we prefer SLS over SA even though there is a gain of up to 5\%, is that (1) the fundamental algorithmic improvement of SA -- the ability to escape from local minima -- is not used in practice, as we will see in subsection \ref{sec:minima}, and (2) tuning the SA parameters is expensive. Several methods have been suggested to tune the initial temperature, such as \cite{Miki2003} and \cite{Ben-Ameur2004}, but they often take several hundred iterations to obtain reliable values (which is quite expensive in our case). The small benefit of SA can be obtained by three other ways. First, by performing SLS runs in parallel (see section \ref{sec:neighb}), second, by increasing the number of iterations, and third by selecting an initial temperature based on previous information such as figure \ref{fig:tmax}. It is likely that a small improvement is obtained without increasing the run time.


\subsection{\label{sec:neighb}Neighborhood structure}
The main parameter of SLS is the neighborhood structure. Choosing an appropriate neighborhood structure is crucial, since it determines the shape of the search space and thus influences the search performance. In \cite{Tian1999} it is observed that the neighborhood structure can have a significant impact on the quality of the solutions for the Traveling Salesman Problem, the Quadratic Assignment Problem, and the Flow-shop Scheduling Problem. 

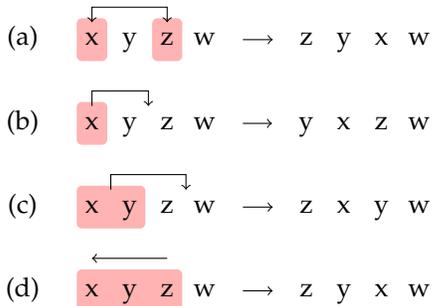
\begin{figure}
\centering
\begin{tikzpicture}[node distance=0.5cm,text height=\heightof{A},text depth=\depthof{y},inner sep=1.5] 
\newcommand*{\tfbox}[1]{{\fboxsep-\fboxrule\fbox{#1}}}

\matrix (m) [column sep=1.0cm] {
\node {(a)};
&[-0.5cm]
\node (x4) {x};\node (y) [right of=x4]{y};\node (z4) [right of=y] {z};\node (w4) [right of=z4]{w};
\draw[<->] (x4.north) -- ($(x4.north) + (0,0.2)$) -- node [midway,above] {} ($(z4.north) + (0,0.2)$) -- (z4.north);
&
\node (x5) {z};\node (y) [right of=x5]{y};\node (z) [right of=y] {x};\node (w5) [right of=z]{w};
\\
\node {(b)};
&
\node (x3) {x};\node (y3) [right of=x3]{y};\node (z3) [right of=y3] {z};\node (w3) [right of=z3]{w};
\draw[->] (x3.north) -- ($(x3.north) + (0,0.2)$) -- node [midway,above] {} ($(z3.north) !.5! (y3.north) + (0,0.2)$) -- ($(z3.north) !.5! (y3.north)$);
&
\node (x6) {y}; \node (y) [right of=x6]{x}; \node (z) [right of=y] {z};\node (w) [right of=z]{w};
\\
\node {(c)};
&]

\node (x1) {x};\node (y1) [right of=x1]{y};\node (z1) [right of=y1] {z};\node (w1) [right of=z1]{w};
\draw[->] ($(x1.north) !.5! (y1.north)$) -- ($(x1.north) !.5! (y1.north) + (0,0.2)$) -- node [midway,above] {} ($(z1.north) !.5! (w1.north) + (0,0.2)$) -- ($(z1.north) !.5! (w1.north)$);
&
\node (x7) {z};\node (y) [right of=x7]{x};\node (z) [right of=y] {y};\node (w) [right of=z]{w};
\\
\node {(d)};
&
\node (x2) {x};\node (y2) [right of=x2]{y};\node (z2) [right of=y2] {z};\node (w2) [right of=z]{w};
\draw[<-] ($(x2.north) + (0,0.2)$) -- node [midway,above] {} ($(z2.north) + (0,0.2)$);
&
\node (x8) {z};\node (y) [right of=x8]{y}; \node (y) [right of=y] {x};\node (w) [right of=z]{w};
\\
};

\tikzstyle{ar}=[shorten >=0.3cm, shorten <=0.3cm, ->]
\draw[ar] (w4) edge (x5);
\draw[ar] (w3) edge (x6);
\draw[ar] (w1) edge (x7);
\draw[ar] (w2) edge (x8);

\begin{scope}[on background layer] 
\node [fill=red!30,anchor=base,rounded corners=2pt, fit=(x4)] {};
\node [fill=red!30,anchor=base,rounded corners=2pt, fit=(z4)] {};
\node [fill=red!30,anchor=base,rounded corners=2pt, fit=(x3)] {};
\node [fill=red!30,anchor=base,rounded corners=2pt, fit=(x1) (y1)] {};
\node [fill=red!30,anchor=base,rounded corners=2pt, fit=(x2) (y2) (z2)] {};
\end{scope}

\end{tikzpicture}
\caption{\label{fig:moves}The elementary neighborhood structures we use. From top to bottom: (a) a single swap, (b) a single shift, (c) shift of a sublist, and (d) mirroring.}
\end{figure}

There are many neighborhood structures for permutation problems such as Horner schemes. For example, a transition could swap two variables in the Horner scheme or move a variable in the scheme. However, there are also neighborhood structures that involve changing larger structures. Figure \ref{fig:moves} gives an overview of four basic transitions from which others can be constructed. From top to bottom, it shows (a) a single swap of two variables in the scheme, (b) a shift of a variable, (c) a shift of a sublist, and (d) a mirroring of a sublist. At each iteration of SLS, a transition to a randomly chosen neighbor is proposed. For the single swap transition, this involves the selection of two random variables in the scheme. 

To examine which neighborhood structure performs best for Horner schemes, we investigate seven (combinations of) neighborhood structures, viz. (1) a single swap, (2) two consecutive swaps, (3) three consecutive swaps, (4) a shift of a single variable, (5) mirroring of a sublist, (6) a sublist shift (which we call `many shift'), and (7) mirroring and/or shifting with an equal probability (which we call `mirror shift'). Swapping multiple times in succession allows for faster traversal of the state space, but also runs the risk to miss states. Moreover, we have tested hybrid transitions, such as performing two consecutive swaps in the first half of the simulation and resorting to single swaps for the latter half, but we found that these combinations did not perform better. In order to present clear plots, we have omitted the plots resulting from these combinations.

\begin{figure}
\centering
\includegraphics[scale=0.4]{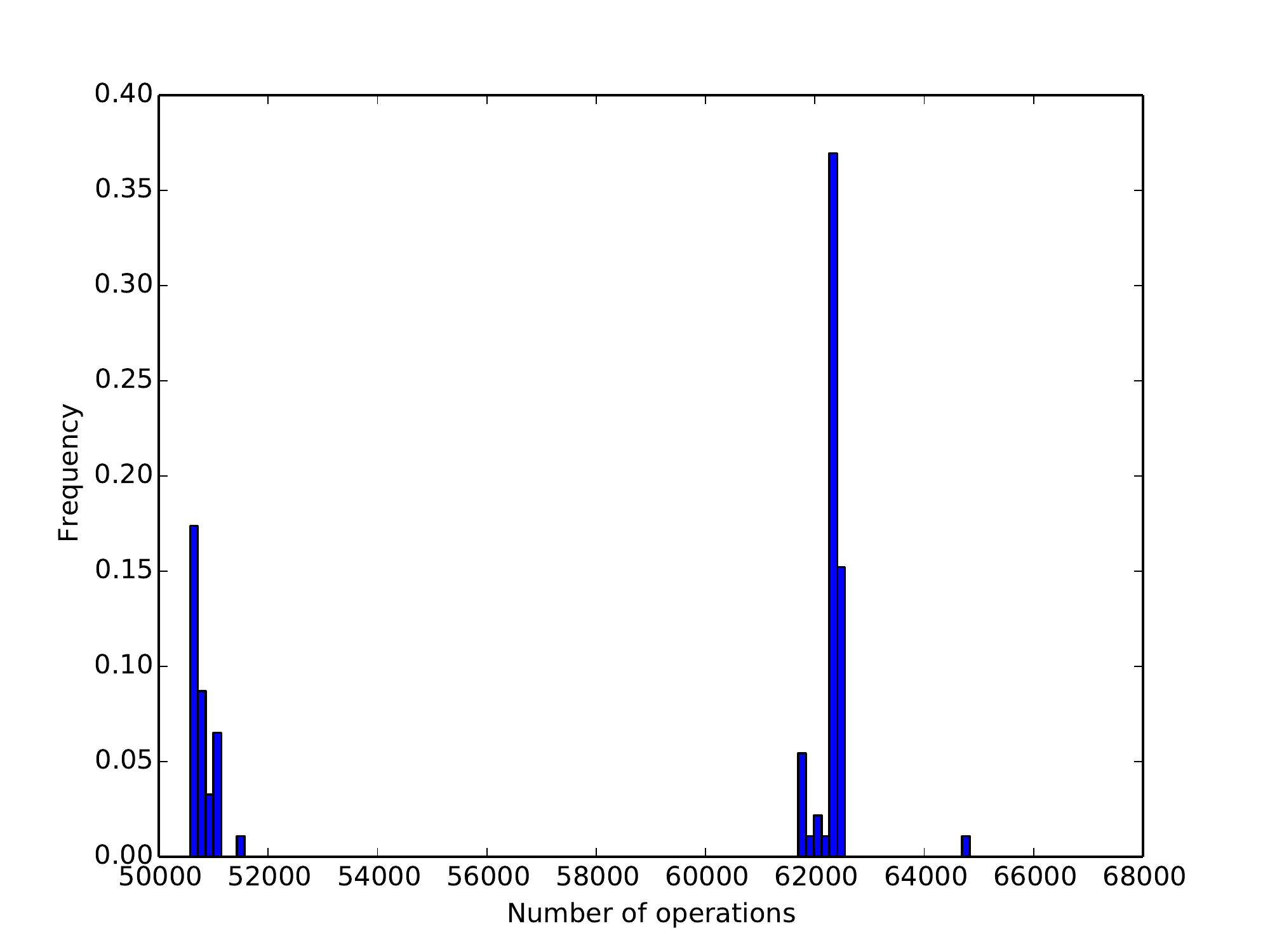}
\caption{\label{fig:F13_1swap} HEP($F_{13}$) expression with 10\,000 runs and 1 swap as a neighborhood structure. Typical for our domain is that there are often two or more spikes. If the simulation is run multiple times, the probability of finding a value close to the minimum is high.}
\end{figure}

We now start investigating two methods of measuring the quality of a neighborhood structure by: (1) the average number of operations obtained by using a neighborhood structure, and (2) the lowest number of operations after performing \emph{several} runs. Figure \ref{fig:F13_1swap} shows the distribution of the number of operations of the expression HEP($F_{13}$) after 10\,000 SLS runs with the neighborhood structure that exchanges two random variables. The average of this distribution is somewhere in the middle, but the actual values that one will measure will be either near 51\,000 or near 62\,000. Thus, the average is not an appropriate measure.

So, we decided to measure the lowest score of several runs, because in practice SLS is run in parallel, and so the results are more in line with those from practical applications.
Thus, we are interested in the neighborhood structure that has the lowest expected value of the minimum of $k$ measurements. Here, we can use the expected value $\mathrm{E}\left[\min\left(X_0,\ldots,X_{k-1} \right)\right]$:
\begin{equation}
V_0 + \sum_{t=0}^{L - 2} (V_{t+1} - V_t) \left(1 - \mathrm{cdf}(D, t)\right)^k
\label{eq:minexp}
\end{equation}
where $k$ is the number of measurements, $X_n$ is the score of the $n$th measurement, $t$ is an index in the discrete distribution, $V_t$ is the number of operations at $t$, $D_t$ is the probability of outcome $V_t$, $L$ is the number of possible outcomes, and cdf the cumulative distribution function. For a derivation, we refer to appendix \ref{sec:minexp}. We shall denote the expected value of the minimum of $k$ runs by $\mathrm{E}_{\text{min},k}$.

Because the number of measurements $k$ is in the exponent in eq. \eqref{eq:minexp}, $\mathrm{E}_{\text{min},k}$ decreases exponentially with $k$ and finally converges to $V_0$ (see figure \ref{fig:minexp_die} in appendix \ref{sec:minexp}). As a consequence, neighborhood structures with a high standard deviation are more likely to achieve better results, since at high $k$ the probability of finding a low value at least once is high. We find that four parallel runs ($k=4$) yield good results.

We will now present detailed results for res(7,6) and HEP($\sigma$) in subsection \ref{sec:res_res}, and for HEP($F_{13}$) and HEP($b$) in subsection \ref{sec:res_hep}. The results for res(7,4), res(7,5), and HEP($F_{24})$ are similar, and are omitted for brevity. res(9,8) is too time consuming for such a detailed analysis (it would take around 35 days to collect all data).

\subsubsection{\label{sec:res_res}Results for res(7,6) and HEP($\sigma$)}

\begin{figure}
\centering
\includegraphics[scale=0.4]{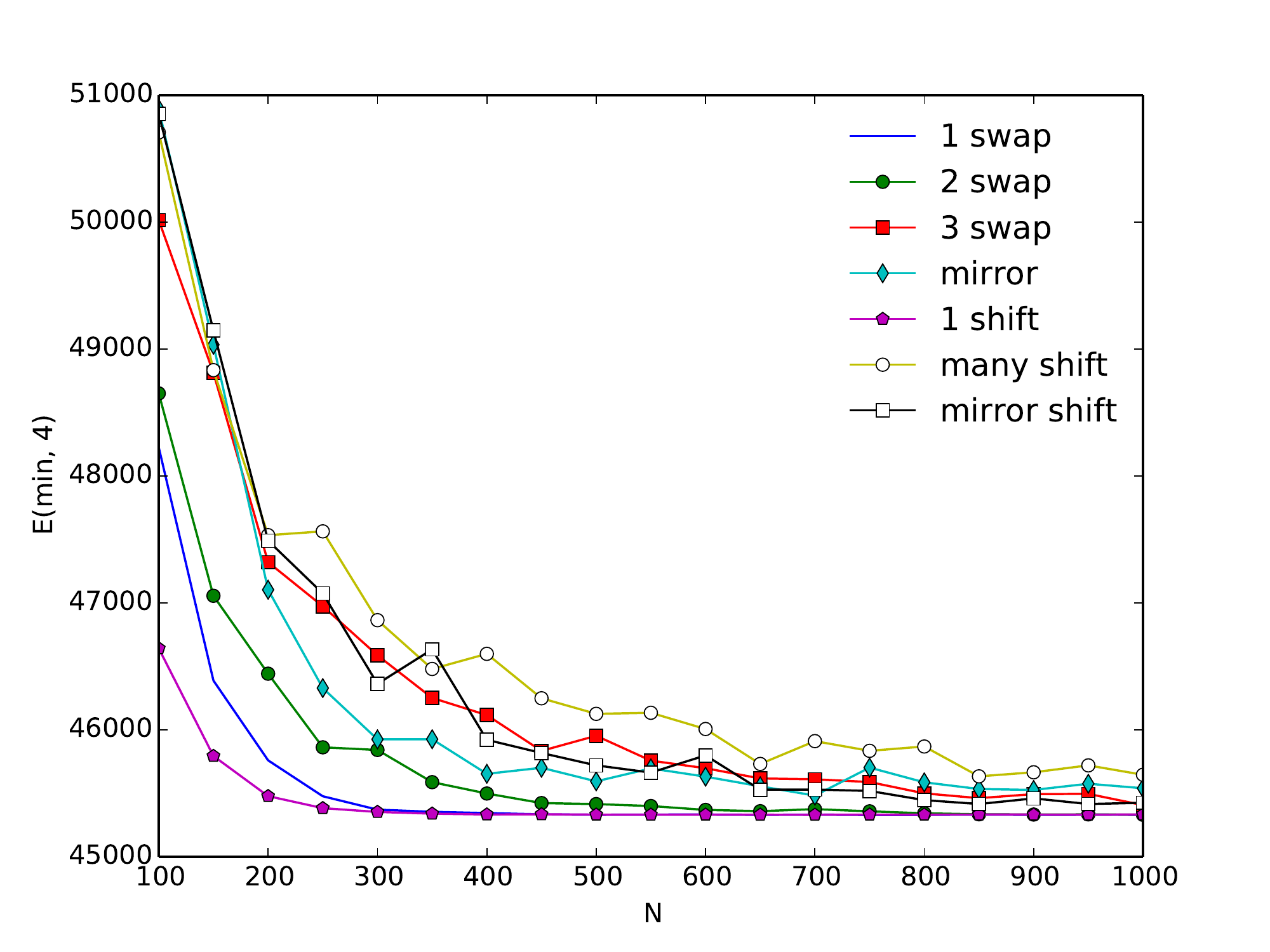}
\caption{The expected number of operations of the minimum of four SLS runs with $N$ iterations for the res(7,6) expression. The 1 shift performs best for low $N$, followed by the 2 swap. All the neighborhood structures converge at $N=900$.}
\label{fig:res76_neighb}
\end{figure}

In figure \ref{fig:res76_neighb} the performance of the neighborhood structures for the expression res(7,6) is shown. We see that shifting a single variable (`1 shift') has the best performance at a low number of iterations $N$, followed by 2 consecutive swaps (`2 swap'). At around $N=900$ all neighborhood structures have converged. Thus, from $N=900$ onward it does not matter which structure is chosen.  

\begin{figure}
\centering
\includegraphics[scale=0.4]{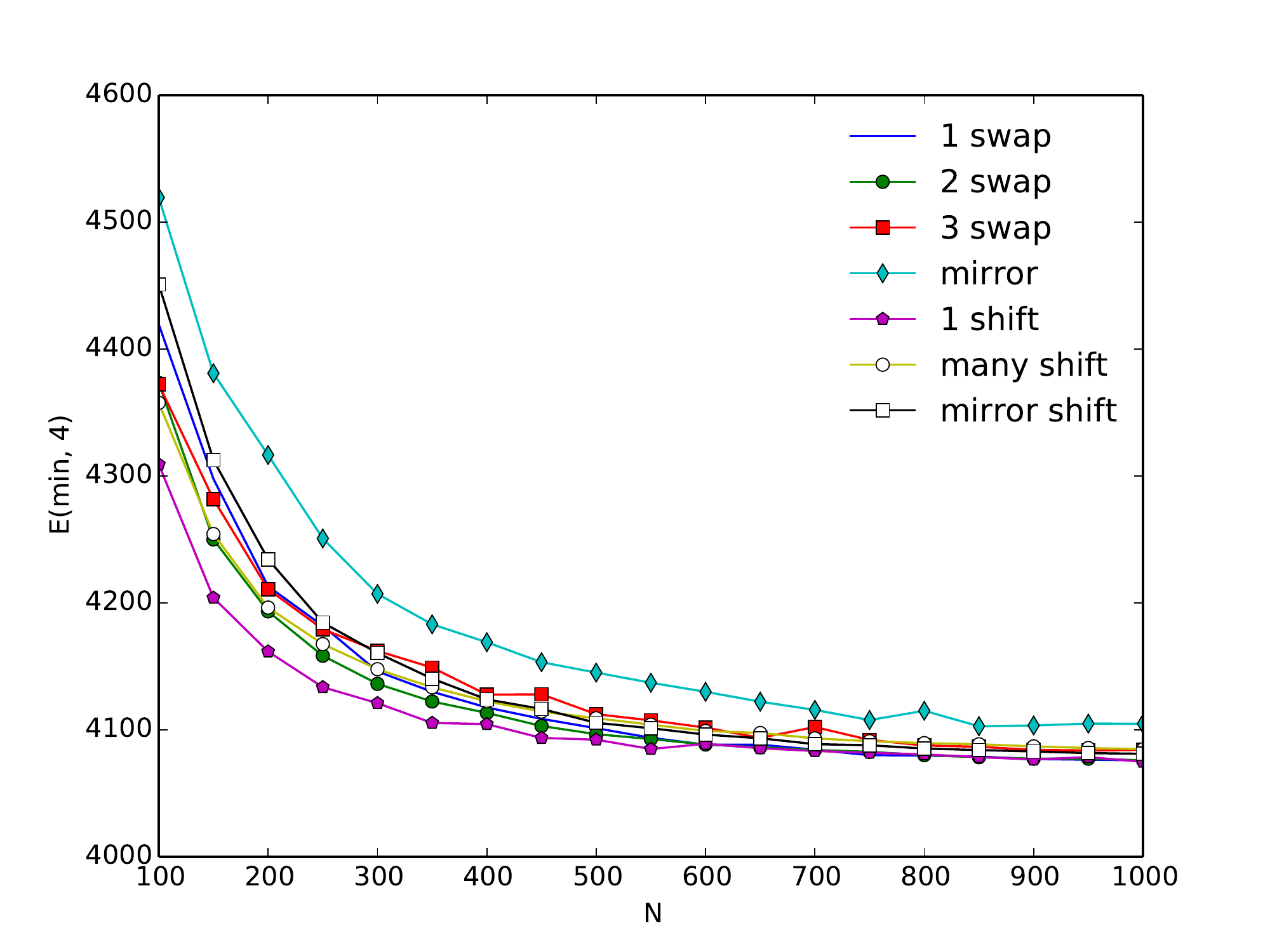}
\caption{The expected number of operations of the minimum of four SLS runs with $N$ iterations for the HEP($\sigma$) expression. The 1 shift performs best for low $N$. All the neighborhood structures converge at $N=600$.}
\label{fig:sigma_neighb}
\end{figure}

In figure \ref{fig:sigma_neighb} we show the performance of the neighborhood structures for HEP($\sigma$). We see that `1 shift' has the best performance at low $N$. At $N=600$, all the neighborhood structures have converged. The characteristics of this plot are similar to those of res(7,6).

We suspect that for a small state space, i.e., a small number of variables, there is not much difference between the neighborhood structures, since the convergence occurs quite early (below $N=1000$). Therefore, we look at two expressions with more variables: HEP($F_{13}$) with 24 variables, and HEP($b$) with 107 variables. 

\subsubsection{\label{sec:res_hep}Results for HEP($F_{13}$) and HEP($b$)}

\begin{figure}
\centering
\includegraphics[scale=0.4]{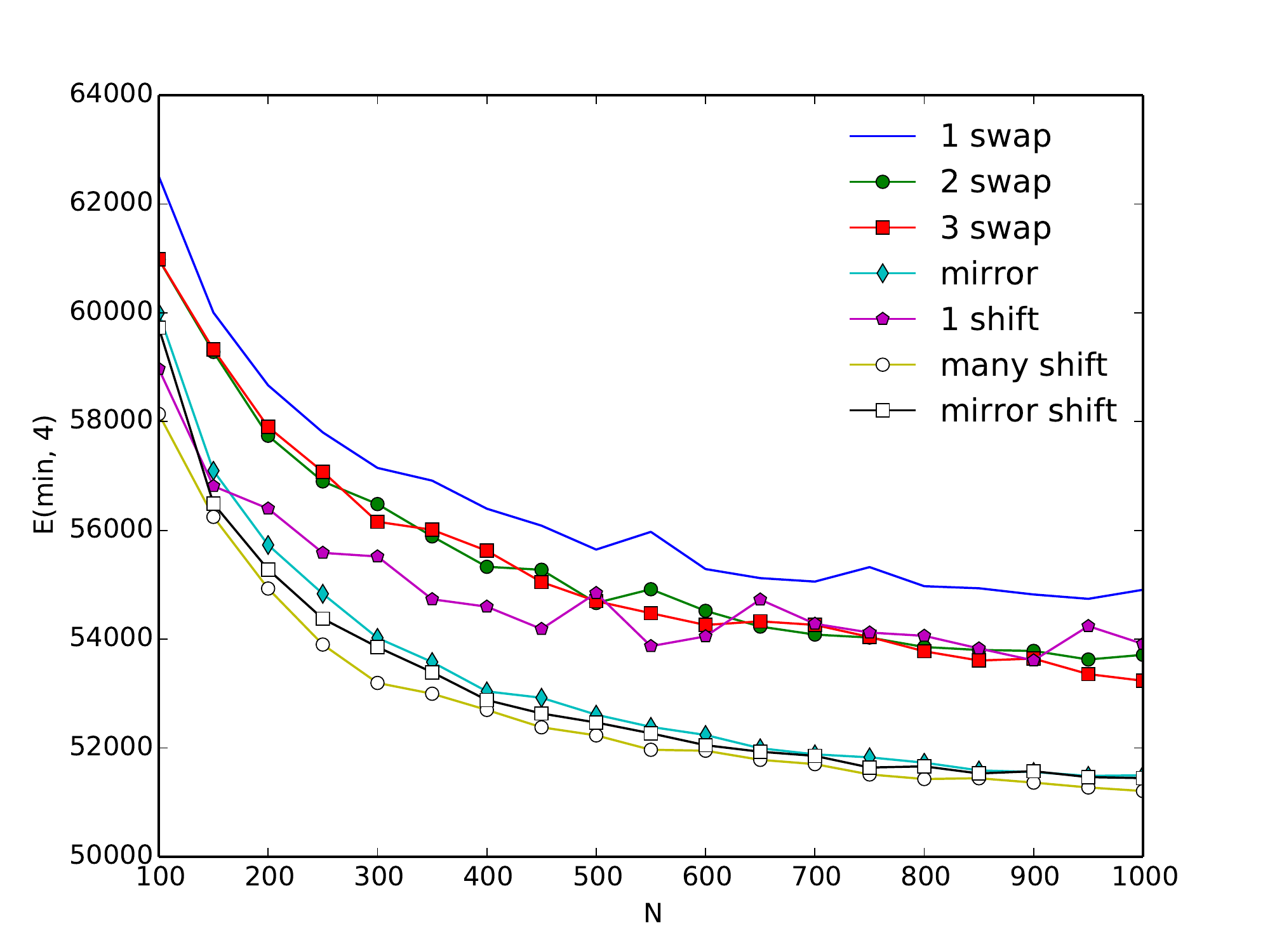}
\caption{The expected number of operations of the minimum of four SLS runs with $N$ iterations for the HEP($F_{13}$) expression. Mirror, many shift, and mirror shift converge to a lower value than the neighborhood structures 1 swap, 2 swap, 3 swap, and 1 shift.}
\label{fig:F13_neighb}
\end{figure}

In figure \ref{fig:F13_neighb} we show the results for HEP($F_{13}$). We see that all the neighborhood structures that involve small changes (`1 swap', `2 swap', `3 swap', and `1 shift') are outperformed by the neighborhood structures that have larger structural changes (`mirror', `many shift', and `mirror shift'). The difference is approximately 8\%. Both groups seem to have converged independently to different values. However, for larger $N$, we expect all neighborhood structures to converge to the same value. The point of convergence has shifted to higher $N$ compared to res(7,6), and HEP($\sigma$), since the state space has increased in size from 15! to 24!. From this plot, we may conclude that the state space of HEP($F_{13}$) is more suited to be traversed with larger changes.

\begin{figure}
\centering
\includegraphics[scale=0.4]{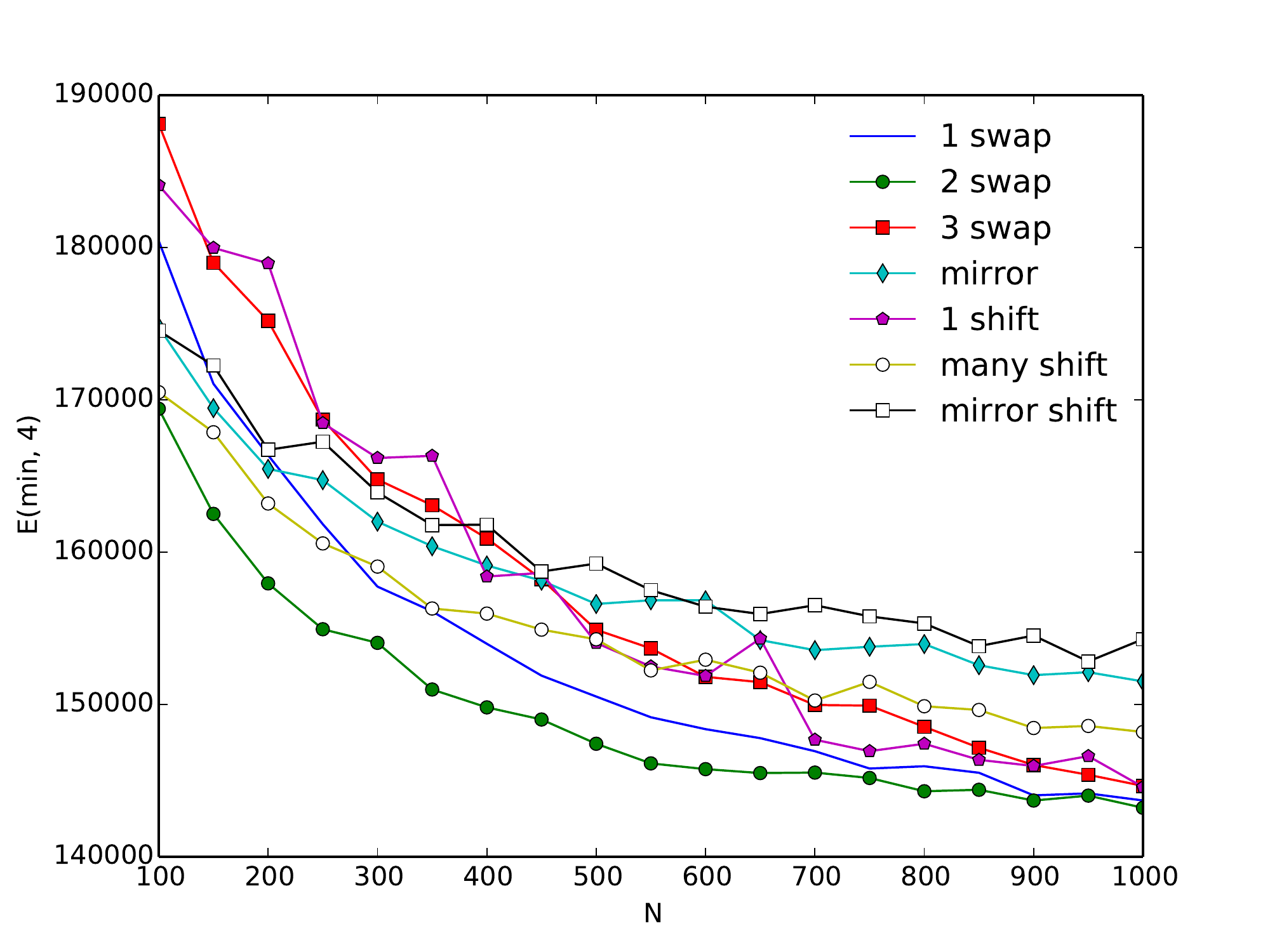}
\caption{The expected number of operations of the minimum of four SLS runs with $N$ iterations for the HEP(b) expression. The 2 swap performs best for low $N$. 1 swap, 2 swap, 3 swap, and 1 shift converge at $N=1000$. The other neighborhood structures perform worse.}
\label{fig:b_neighb}
\end{figure}

In figure \ref{fig:b_neighb} the performance of the neighborhood structures for HEP($b$) is shown. We see that two consecutive swaps perform best at low $N$ and that `1 swap', `2 swap', `3 swap', and `1 shift' converge at $N=1000$. The neighborhood structures that involve larger structural changes (`mirror', `many shift', `mirror shift') perform worse. These results are different from those of HEP($F_{13}$): for HEP($b$), with an even larger state space than HEP($F_{13}$), smaller moves are better suited. This means that the mere number of variables is not a good indicator for the selection of a neighborhood scheme. 

\subsubsection{\label{sec:res_comb}Combined results}

For the four benchmark expressions displayed above, and for the other three benchmark expressions, we observe that the relative improvement of the choice of the best neighborhood structure compared to the worst neighborhood structure is never more than 10\%. Furthermore, we observe that there are two groups of neighborhood structures when the state space is sufficiently large: a group with small changes to the state (`1 swap', `2 swap', `3 swap', `1 shift'), and a group with large structural changes (`mirror', `many shift' `mirror shift'). These two groups converge before $N=1000$ for expressions with small state spaces, such as HEP($\sigma$), but are further apart for expressions with more variables, such as HEP($F_{13}$) and HEP($b$). The difference in quality in the group itself is often negligible (less than 3\%). Thus, as a strategy to apply the appropriate neighborhood structure, we suggest to distribute the number of parallel runs evenly among the two groups: in the case of four runs, two of the runs can be performed using a neighborhood structure from the small change group and two using a structure from the large change group. 


\subsection{\label{sec:prop}State space properties}
The fact that SLS works so well is surprising. Two well-known obstacles are (1) a local search can get stuck in local minima which yields inferior results, and (2) the Horner scheme problem does not have local heuristics, so there is no guidance for any best-first search. Remarkably, SLS only needs 1000 iterations for a 107 variable expression (HEP($b$)) to obtain good results, whereas a TSP benchmark problem with a comparable state space size, viz. kroA100 \cite{TSPLib2008} with 100 variables, takes more than a million iterations to converge using a manually tuned SA search.

A thousand iterations is also a small number compared to the size of the state space. The average distance between two arbitrary states is 98 swaps. A thousand iteration SLS search accepts approximately $300$ suggested swaps, so at least 33\% of all the accepted moves should move towards the global minimum. This scenario would be unlikely if the state space is unsuited for SLS, so perhaps the state space has convenient properties for our purposes. At this moment we consider the following two conditions: local minima are rare, and the region of the global minimum is flat. We discuss the two conditions in the following subsections.

\subsubsection{\label{sec:minima}Local minima and saddle points}

\begin{figure*}
\centering
\begin{subfigure}[b]{0.325\linewidth}
\includegraphics[width=\linewidth]{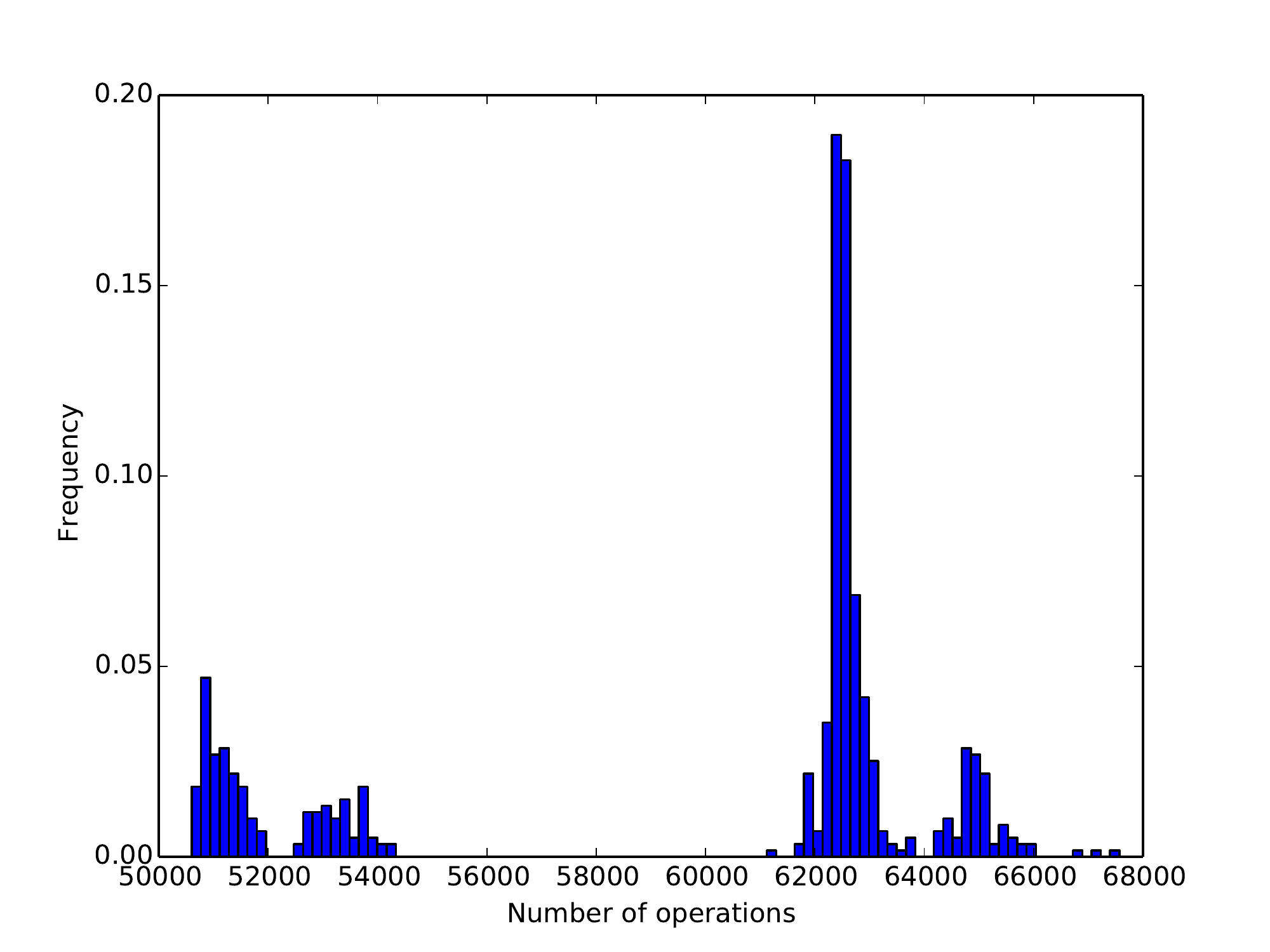}
\end{subfigure}
\begin{subfigure}[b]{0.325\linewidth}
\includegraphics[width=\linewidth]{data/F13_10k_1swap.pdf}
\end{subfigure}
\begin{subfigure}[b]{0.325\linewidth}
\includegraphics[width=\linewidth]{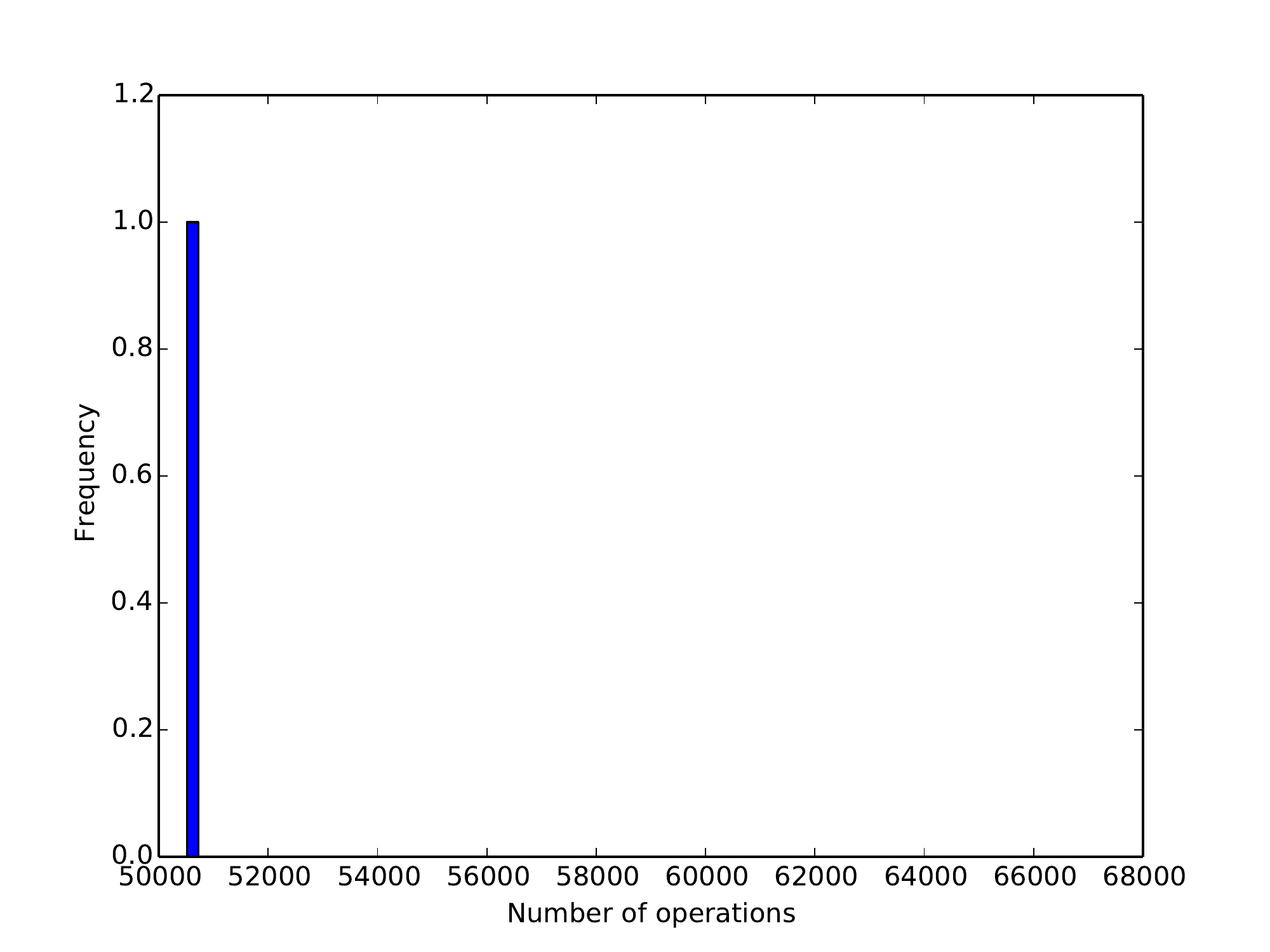}
\end{subfigure}
\caption{The distribution of the number of operations for the HEP($F_{13}$) expression with 1 swaps for 1000 iterations (left), 10\,000 iterations (middle) and 100\,000 iterations (right) at $T=0$. There appears to be a local minimum around 62\,000. However, as the number of iterations is increased, the local minimum becomes smaller relative to the global minimum region (middle) and completely disappears (right). We may conclude that the apparent local minimum is not a local minimum, but a saddle point, since SLS is able to escape from the `minimum'.}
\label{fig:F13_minima}
\end{figure*}

To obtain an idea on the number of local minima, we measure how often the simulation gets stuck: if there are many local minima, we expect the simulation to get stuck often. In figure \ref{fig:F13_minima}, we show the distribution of HEP($F_{13}$) for $1000$, $10\,000$, and $100\,000$ SLS runs respectively. For $1000$ and $10\,000$ runs we see two peaks: one at the global minimum near $51\,000$ and one at an apparent local minimum near $62\,000$. As the number of iterations is increased, the weight shifts from the apparent local minimum to the global minimum: at 1000 iterations, there is a probability of 27.5\% of arriving in the region of the global minimum, whereas this is $36.25\%$ at $10\,000$ iterations. Apparently the local minimum is `leaking': given sufficient time, the search is able to escape. The figure on the right with 100\,000 iterations confirms the escaping possibility: the apparent local minimum has completely disappeared. Thus, the local minimum is in reality a saddle point, since for a true local minimum there is no path with a lower score leading away from the minimum. Since SLS requires many iterations to escape from the saddle point, only a few transitions reduce the number of operations.


We observe that apparent local minima disappear for our other benchmark expressions as well. SLS runs with $100\,000$ iterations approach the global minimum for all of our benchmark expressions. For example, for HEP($F_{13}$) mentioned above, the result is $50636 \pm 57$ and for HEP($\sigma$) the result is $4078 \pm 9$. The small standard deviations indicate that no runs get stuck in local minima (at least not in local minima significantly higher than the standard deviation).

From these results we may conclude that true local minima, from which a local search cannot escape, are rare for Horner schemes.


\subsubsection{\label{sec:flatness}Flatness of the state space}
To build an intuition for what the state space looks like, we consider its flatness. We measure how many of the neighbors have a value (number of operations) that does not differ by more than 1\%: $\frac{|x_n - x|}{|x|} < 1\%$, where $x$ is the reference state and $x_n$ is a neighbor of $x$. For brevity, we shall refer to this as `close'.

\begin{figure}
\centering
\includegraphics[scale=0.4]{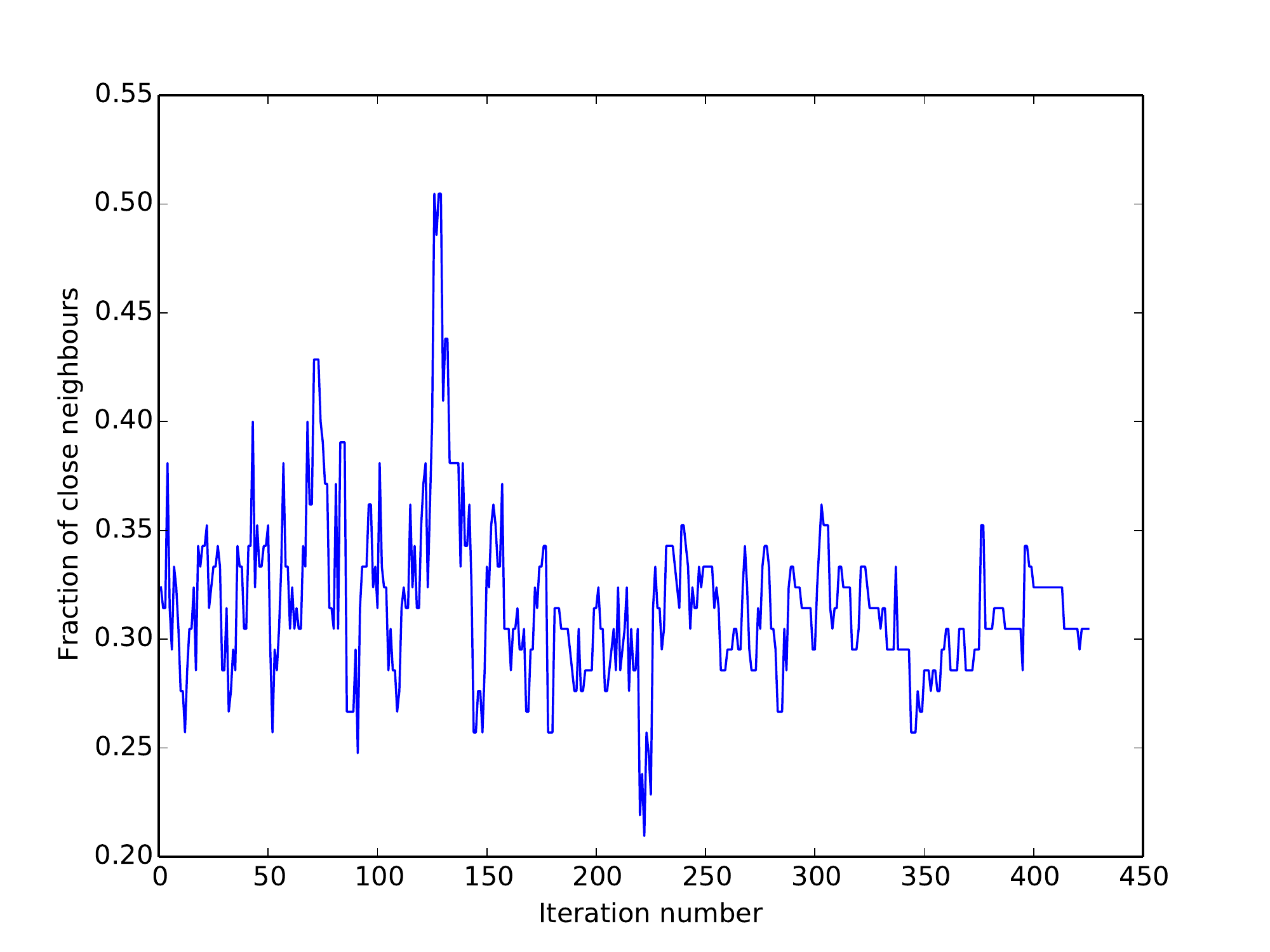}
\caption{The number of close neighbors (value difference within 1\%) for the current state during a typical SLS run for the HEP($\sigma$) expression. Contrary to the TSP kroA100 problem, HEP($\sigma$) does not show a decrease in the number of close neighbors, but remains steady around 30\%. This is an indication that the state space is flat.}\label{fig:sigma_deg}
\end{figure}

\begin{figure}
\centering
\includegraphics[scale=0.4]{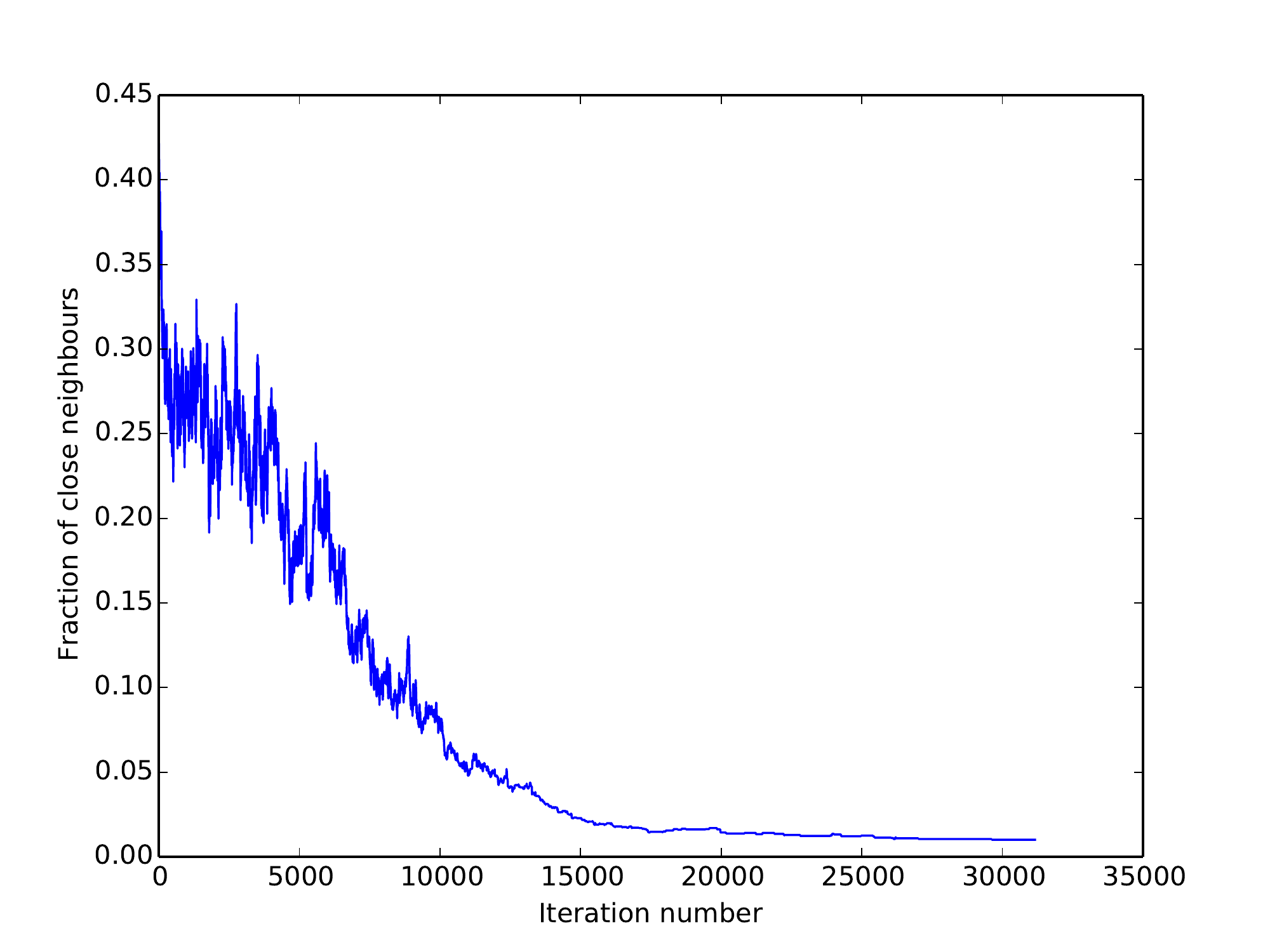}
\caption{The number of close neighbors (value difference within 1\%) for the current state during a typical SA run for the TSP benchmark problem kroA100 with 100 cities. The early states have many close neighbors, but as the simulation is converging to a minimum, the number of close neighbors decreases to about 0.9\%. This is an indication that the state space has steep (local) minima.}\label{fig:tsp_deg}
\end{figure}

In figure \ref{fig:sigma_deg} we show the results for HEP($\sigma$) for the current states during a typical single SLS run. We see that throughout the simulation the percentage of close neighbors is approximately 30\%. We compare these results to an SA run of the TSP problem kroA100 (displayed in figure \ref{fig:tsp_deg}). We see that for a random starting state the number of close neighbors is 30\% as well, but as the simulation approaches the global minimum (at the right of the graph), the number of close neighbors decreases to 0.9\%. As a result, the global minimum for TSP must be very narrow.

\begin{figure}
\centering
\includegraphics[scale=0.4]{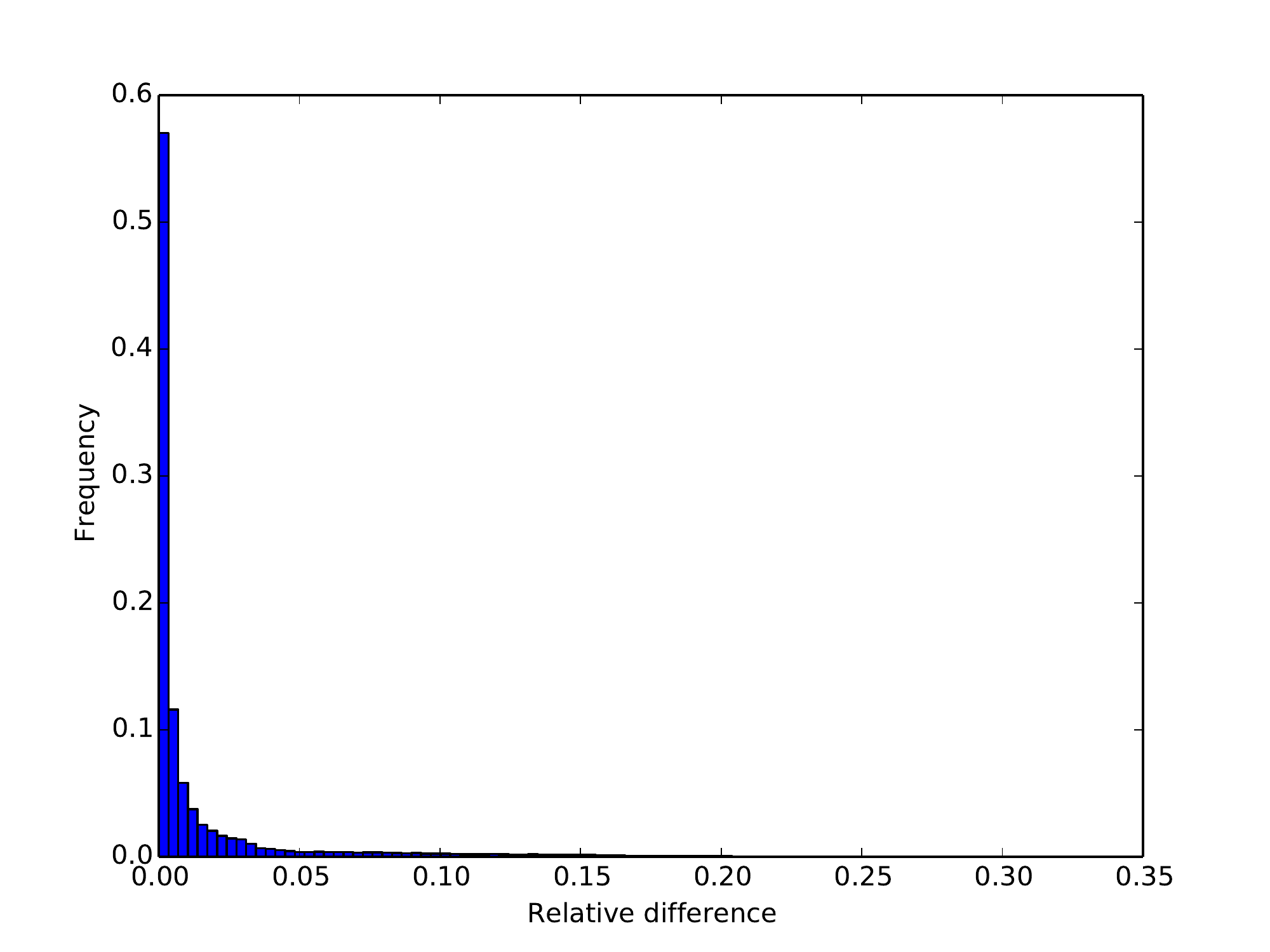}
\caption{Relative difference of the values of swap neighbors of global minima for the HEP($b$) expression, sampled over 37 states. The mean is $0.015 \pm 0.034$. The region is very flat: we observe that 85\% of the neighbors have a value that is within 1\% of the current value.}\label{fig:b_deg}
\end{figure}

These results are a first hint that the state space of Horner is flat and terrace-like, whereas the TSP problem is more trough-like, with steep global/local minima. To investigate the flatness more deeply, we have looked at the distribution of the relative difference $\frac{|x_n - x|}{|x|}$. For the global minimum of the HEP($b$) expression, this is depicted in figure \ref{fig:b_deg}. We see that about 75\% of the neighbors are within 1\% and 95\% within 5\%, which is even higher than for HEP($\sigma$). We observe similar features for the other points in the state space, including hard to escape saddle points. 

The property that the state space is flat is not only present in physics expressions, but is found in our other four benchmark expressions as well. Additionally, we have generated test expressions that we know to have interesting mathematical structures, such as powers of expressions. For example, for the expression $(4a+9b+12c^2+2d+4e^3-2f+8g^2-10h+i-j+2k^2-3j^4+l-15m^2)^6$, 43\% of the neighbors, $18$\% of the second neighbors and 7.3\% of the third neighbors are close.

The question arises why the number of close neighbors is so high for the HEP($b$) expression. For most expressions it is around 30\%, but for HEP($b$) it is 75\%. A closer inspection revealed that the HEP($b$) expression has the special property that 90 of the 107 variables never appear in the same term: a term that contain variable $x$ does not contain variable $y$ and vice versa. As a result, the Horner schemes $x,y$ and $y,x$ yield the same expression. The HEP($b$) expression is not alone in this property: it represents a class of problems that often appears in electron-positron scattering processes. 

The fact that some variables do not appear together in the same term is caused by a symmetry of the expression, since rearranging these variables in the scheme does nothing if they are direct neighbors in the scheme. The more symmetrical the expression is, the more likely it is that neighbors have the exact same value or a close value (within 5\%). In the case of a uniformly random expression where the number of terms is much greater than the number of variables, we expect that practically all swaps are ineffective. The reason is that there is a high probability that each variable appears in an equal number of terms and has equal mixing. 


Many, if not all, large expressions exhibit the `flatness' property of their state space, since in most cases the number of terms is much larger than the number of variables. For the expressions that we have tested, the ratio of the number of terms and the number of variables is always more than a factor $1000$. As a consequence, most variables will appear in many terms, which in turn increases uniformity, resulting in neighboring states with small differences in value (less than 5\%). 

\subsection{\label{sec:performance}Performance of SLS vs MCTS}
Below, we compare the results of Stochastic Local Search to the previous best results from MCTS, for our eight benchmark expressions res(7,4), res(7,5), res(7,6), res(9,8), HEP($\sigma$), HEP($F_{13}$), HEP($F_{24}$) and HEP($b$). The results of all the MCTS runs except for res(9,8), and HEP($b$) are taken from \cite{Kuipers2013}.\footnote{We only consider optimizations by Horner schemes and CSEE. Additional optimizations that are mentioned in \cite{Kuipers2013}, such as `greedy' optimizations, can just as well be applied to the results of SLS.} The results are displayed in table \ref{tbl:perf}.


\begin{table*}[t]
\centering
\begin{tabular}{|l|c|c||c|c||c|c|c|}
\cline{2-8}
\multicolumn{1}{c|}{} & vars & original & MCTS 1k & MCTS 10k & SLS 1k & SLS 10k & $\mathrm{E}_{\text{min},4}$ 1k\\
\hline
res(7,4)    & 13    &   29\,163 & $(3.86 \pm 0.1)\cdot 10^3$ & $(3.84 \pm 0.01)\cdot 10^3$ & $(3.92 \pm 0.28)\cdot 10^3$ & $3834 \pm 26$ & $3819 \pm 9$\\
res(7,5)    & 14      &  142\,711 & $(1.39 \pm 0.01)\cdot 10^4$ & $13768 \pm 28$ & $13841 \pm 441$ & $13767 \pm 21$ & $13770 \pm 5$\\
res(7,6)    & 15      &  587\,880 & $(4.58 \pm 0.05)\cdot10^4$ & $(4.54 \pm 0.01)\cdot10^4$ & $46642 \pm 3852$ & $(4.61 \pm 0.25) \cdot 10^4$ & $(4.55 \pm 0.16) \cdot 10^4$\\
res(9,8)    & 19      &  83\,778\,591 & $(5.27 \pm 0.25)\cdot10^6$ & $(4.33 \pm 0.31)\cdot10^6$ & $(4.13 \pm 0.34) \cdot 10^6$ & $(4.03 \pm 0.17) \cdot 10^6$ & $(3.97 \pm 0.18) \cdot 10^6$\\
\hline
HEP($\sigma$) & 15 & 47\,424 & $4114 \pm 14$ & $4 087 \pm 5$ & $4226 \pm 257$ & $4082 \pm 58$ & $4075 \pm 25$\\
HEP($F_{13}$) & 24 & 1\,068\,153 & $(6.6 \pm 0.2)\cdot 10^4$ & $(6.47\pm 0.08)\cdot 10^4$ & $(5.99 \pm 0.51)\cdot 10^4$ & $(5.80 \pm 0.55) \cdot 10^4$ & $(5.37 \pm 0.40) \cdot 10^4$\\
HEP($F_{24}$) & 31 &7\,722\,027 & $(3.80 \pm 0.06)\cdot 10^5$ & $(3.19 \pm 0.04)\cdot 10^5$ & $(3.16 \pm 0.23)\cdot 10^5$& $(3.06 \pm 0.23) \cdot 10^5$ & $(2.98 \pm 0.09) \cdot 10^5$ \\
HEP($b$) & 107 &1\,817\,520 & $(1.81 \pm 0.04) \cdot 10^5$ & $(1.65 \pm 0.08) \cdot 10^5$ & $(1.50 \pm 0.08)\cdot 10^5$ & $(1.40 \pm 0.06)\cdot 10^5$ & $(1.44 \pm 0.04) \cdot 10^5$\\
\hline
\end{tabular}
\caption{\label{tbl:perf} SLS compared to MCTS. The MCTS results for all expressions except res(9,8) and HEP($b$) are from \cite{Kuipers2013}. All the values are statistical averages over at least 100 runs. SLS results have a larger standard deviation, and thus the expected value of the minimum is often lower than these values (see last column).}
\end{table*}

The results for MCTS with 1000 and 10\,000 iterations are obtained after considerable tuning of $C_p$ and after selecting whether the scheme should be constructed forward or in reverse (i.e., the scheme is applied backwards \cite{Ruijl2014}). 

For smaller problems, we observe the seemingly negative result that the averages of SLS are on a par with or slightly worse than MCTS. However, as a positive result we see that the standard deviations of SLS are higher than MCTS. Consequently, we expect SLS to outperform MCTS if several runs are performed in parallel. Indeed, this is what we see in the last column of table \ref{tbl:perf}. The standard deviations of MCTS are often an order of magnitude smaller than those of SLS, so the benefits of running MCTS in parallel are smaller. We may conclude that although for some small problems MCTS has better average scores, SLS has better minimal behavior if it is run in parallel.

For our largest expressions, HEP($F_{13}$), HEP($F_{24}$) and HEP($b$), we observe that SLS with $1000$ iterations yields better results than MCTS with $10\,000$ iterations. For HEP($F_{24}$), the average of SLS with 1000 iterations is about 20\% better than the average for MCTS with 1000 iterations. In fact, the results are slightly better than MCTS with $10\,000$ iterations. If we take the $\mathrm{E}_{\text{min},4}$ into account, the expected value for HEP($F_{24}$) is an additional 7\% less.

The fact that SLS outperforms MCTS when the number of variables is greater than $23$, may be due to the fact that there are not sufficient iterations for the branches to reach the bottom, making the choice of the last variables essentially random (see section \ref{sec:motivation} and \cite{Ruijl2014}). This may also be the reason why for MCTS it is important whether the scheme is constructed forward or in reverse: if most of the performance can be gained by carefully selecting the last variables, building the scheme in reverse will yield better performance.

SLS is 10 times faster (in clock time) than MCTS, since most of the time is spent in the evaluation function. It is able to make reductions ranging from a factor 7 for our smallest expression, to 26 for our largest expression. The reduction factor becomes larger when there are more operations.


\section{\label{sec:conclusion}Conclusions}

Throughout the last forty years many algorithms have been devised to solve hard optimization problems. Each new algorithm introduces complexity in the form of parameters that have to be fine-tuned manually to the problem. It is tempting to use the latest successful and complex algorithms, but in doing so, sometimes convenient properties of the problem class can be missed or can remain unused. An analysis of the problem space can help identify which algorithm is best suited. In the case of Horner schemes, we have found that one of the most basic algorithms, Stochastic Local Search, yields the best results.  

Stochastic Local Search provides a search method with two parameters: the number of iterations (computation time) and the neighborhood structure. We find that running half of the simulations with a neighborhood structure that makes minor changes to the state (i.e., a single shift of a variable), and running the other half with a neighborhood structure that involves larger changes (i.e., the mirroring of a random sublist) is a good strategy for all of our benchmark expressions (see subsection \ref{sec:res_comb}). Consequently, only the computation time remains as a parameter. We find that (1) SLS obtains similar results to MCTS for expressions with around 15 variables, (2) SLS outperforms MCTS for expressions with 24 or more variables, and (3) SLS requires ten times fewer samples than MCTS to obtain similar results. Therefore we may conclude that SLS is more than 10 times faster.

The result that a basic algorithm such as SLS performs well is surprising, since Horner schemes have some properties that make the search hard: there are no known local heuristics, and evaluations could take several seconds. In the previous sections we have shown that the performance of SLS is so good because the state space of Horner schemes is flat and has few local minima.

The number of operations is linearly related to the time it takes to perform numerical evaluations. The difference between the number of operations for the unoptimized and the optimized expression is more than a factor 24 compared. As a consequence, we are able to perform numerical integration (via repeated numerical evaluations) at least 24 times faster.

For High Energy Physics, the contribution is immediate: numerical integration of processes that are currently experimentally verified at CERN can be done significantly faster. Additionally, expensive calculations that would have taken months for the next-generation particle collider ILC can be done in days or even hours.

Our algorithms will be implemented in the next release of the open source symbolic manipulation system FORM \cite{Vermaseren2014}.

\section{\label{sec:discussion}Discussion / future work}
Currently, our algorithms assume that the expressions are commutative, but our implementation could be expanded to be applied to generic expressions with non-commuting variables. Especially in physics, where tensors are common objects, this is useful. Horner's rule can only be applied uniquely to commutative variables, but the pulling outside brackets keeps the order of the non-commuting objects intact. Thus, for Horner's rule the only required change is the selection of commutative variables for the scheme. The common subexpression elimination should honor the ordering of the non-commutative objects. For example, in figure \ref{fig:cse}, the two highlighted parts are not a common subexpression if the variables are non-commutative ($a+e$ cannot be moved to the left of $c$). To enable non-cummutative objects, CSEE should only compare connected subsets.

Additional work can be put in finding other methods of reductions. For example, expressing certain variables as linear combinations of other variables may reduce the number of operations even further. Many of these patterns cannot be recognized by common subexpression elimination alone. Determining which variables should be expressed as linear combinations of other variables to yield optimal results is an open problem. Perhaps techniques such as Local Stochastic Search are applicable to this subject as well.

\appendices
\section{\label{sec:minexp}Expected value of minimum}
In order to provide a wider accessibility, we provide the following derivation of the expected value of the minimum of $n$ samples, used for eq. \ref{eq:minexp}. 

We draw $n$ numbers $X_0,\ldots, X_{n-1}$ from the discrete probability distribution $D$, where $D_t$ is the probability of outcome $V_t$, $L$ is the number of outcomes, where $t$ is an index ranging from $0$ to $L-1$ and $V_a < V_b$ iff $a < b$ (thus $D$ can be viewed as a histogram). We want to know $E(\min(X_0,\ldots,X_{n-1}))$.
Let:
\begin{align}
\begin{split}
\phi(t) &\equiv P(\min(X_0,\ldots,X_{n-1}) < V_t) = 1 - P(\forall X_i \geq V_t)
\\&= 1 - P(X_i \geq V_t)^n = 1 - (1 - P(X_i < V_t))^n \\&= 1 - (1 - cdf(D, t - 1))^n 
\end{split}
\label{eq:phi}
\end{align}
where cdf is the cumulative distribution function.

We abbreviate $\min(X_0,\dots,X_{n-1})$ to $\min(n)$. 
Using eq. \eqref{eq:phi}, we find an expression for the probability that the minimum is $V_t$:
\begin{align}
\begin{split}
P(\min(n) = V_t) =& P(\min(n) < V_{t+1})- P(n) < V_{t})\\=& \phi(t + 1) - \phi(t)\\
=& (1 - \text{cdf}(D, t - 1))^n - (1 - \text{cdf}(D, t))^n
\end{split}
\end{align}

Finally, the expected value is:
\begin{align}
\begin{split}
E(\min(n) =& \sum_t V_t P(\min(n) = V_t)\\
       =& \sum_{t=0}^{L-1} V_t ((1 - \text{cdf}(D, t - 1))^n \\
       &-(1 - \text{cdf}(D, t))^n)\\
       =& V_0 + \sum_{t=0}^{L-2} (V_{t+1}-V_t) (1 - \text{cdf}(D, t))^n
\end{split}
\end{align}

We see that the expected value decreases exponentially in $n$ to $V_0$.

\begin{figure}
\centering
\includegraphics[scale=0.4]{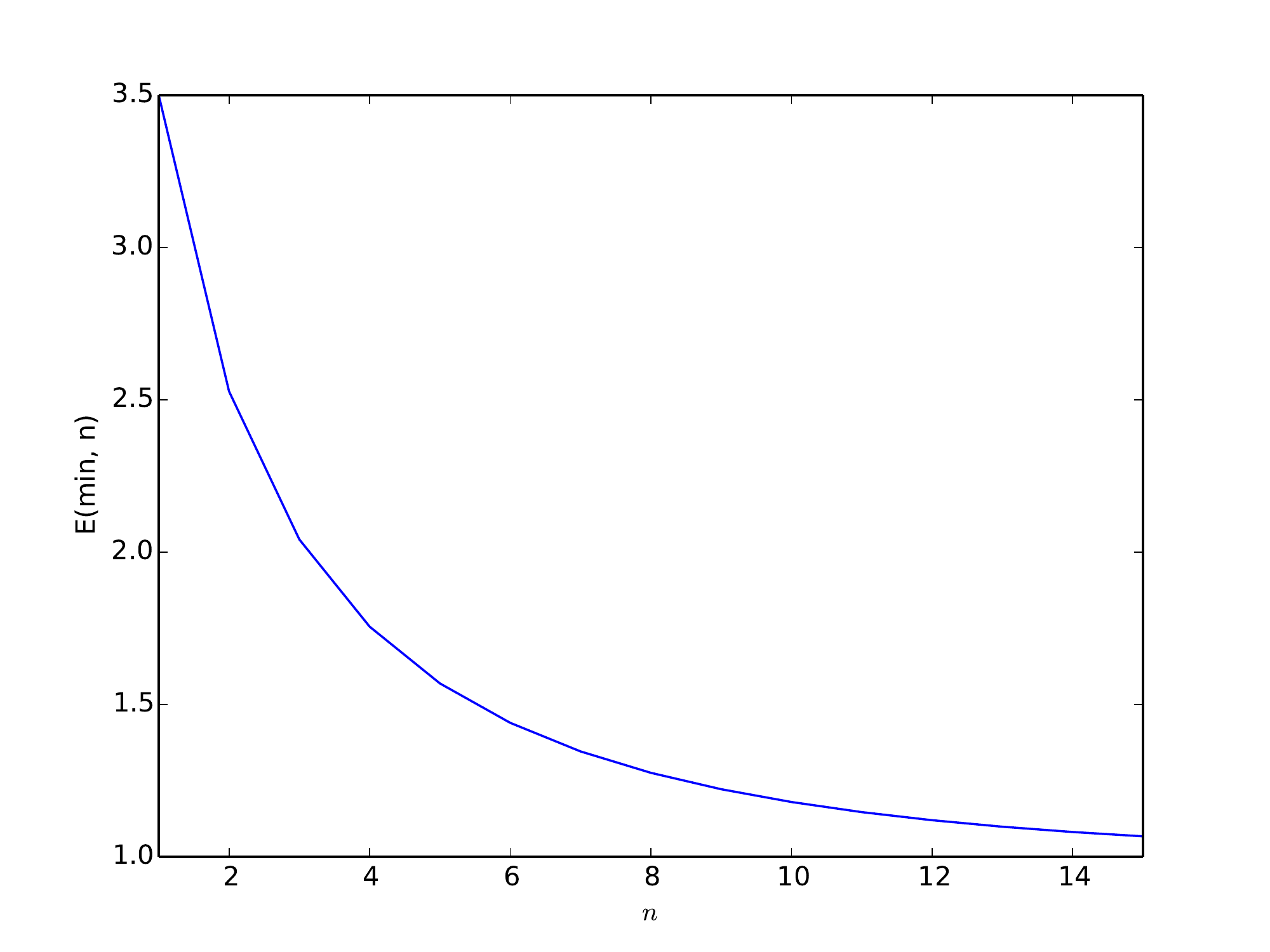}
\caption{\label{fig:minexp_die} The expected value of the minimum of $n$ 6-sided dice throws. This number decreases exponentially from $3.5$ to $1$.}
\end{figure}

Figure \ref{fig:minexp_die} shows the dependence of the expected value of the minimum of $n$ runs for a fair 6-sided die. For $n=1$, the result is the average value $3.5$. $E(\min(X_0,\ldots,X_{n-1}))$ decreases exponentially to $1$.

\ifCLASSOPTIONcompsoc
  \section*{\label{sec:ac}Acknowledgments}
\else
  \section*{\label{sec:ac}Acknowledgment}
\fi

This work is supported in part by the ERC Advanced Grant no. 320651, ``HEPGAME''.

\bibliographystyle{IEEEtran}
\bibliography{IEEEabrv,ref}

\newpage


\begin{IEEEbiography}[{\includegraphics[width=1in,height=1.25in,clip,keepaspectratio]{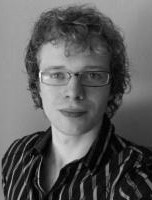}}]{Ben Ruijl}
Ben Ruijl is a Ph.D. researcher at Nikhef Amsterdam and Leiden Institute of Advanced Computer Science of Leiden University. He obtained his masters degree (with honors) in theorerical physics in Nijmegen. His main fields of research are theoretical physics, combinatorial problems, and machine learning. 
\end{IEEEbiography}

\begin{IEEEbiography}[{\includegraphics[width=1in,height=1.25in,clip,keepaspectratio]{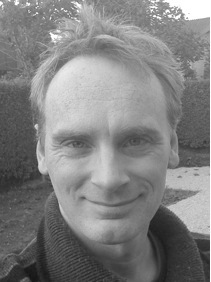}}]{Aske Plaat}
Aske Plaat is professor of Data Science at the Leiden Centre of Data Science of the Leiden Institute of Advanced Computer Science of Leiden University, The Netherlands. His main fields of research are data science, machine learning and artificial intelligence. Many of the questions that he has worked on include large scale algorithms, combinatorial search, affective computing, complexity, strategy, and computer chess, for which he received the Novag Award for the best computer chess publication. Previously, he has worked at the University of Alberta, the Massachusetts Institute of Technology, Vrije Universiteit Amsterdam, Tilburg University and NHTV Breda.
\end{IEEEbiography}


\begin{IEEEbiography}[{\includegraphics[width=1in,height=1.25in,clip,keepaspectratio]{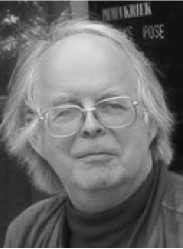}}]{Jos Vermaseren}
Jos Vermaseren obtained his masters degree in Nijmegen (1973) and his Ph.D. 
in theoretical physics at the State University of New York at Stony Brook 
(1977). Since 1981 he is a member of the theory group at Nikhef, the Dutch 
institute for subatomic physics. His field is the phenomenology of particle 
physics and his specialty is higher order calculations in quantum field 
theory. For this he has created the symbolic manipulation program FORM that 
is widely used for complicated algebraic calculations involving very large 
formulas. It was also vital for solving several problems in mathematics. In 
2007 he received the Humboldt award for a number of symbolic calculations 
that had great impact. In 2012 he was awarded an ERC advanced grant for a 
new field of investigations, combining AI game techniques with automatized 
physics calculations (HEPGAME). He is a regular visitor at the Universidad 
Autonoma de Madrid, DESY Zeuthen, the Karlsruhe Institute of Technology and 
the Japanese national laboratory KEK at Tsukuba.
\end{IEEEbiography}

\begin{IEEEbiography}[{\includegraphics[width=1in,height=1.25in,clip,keepaspectratio]{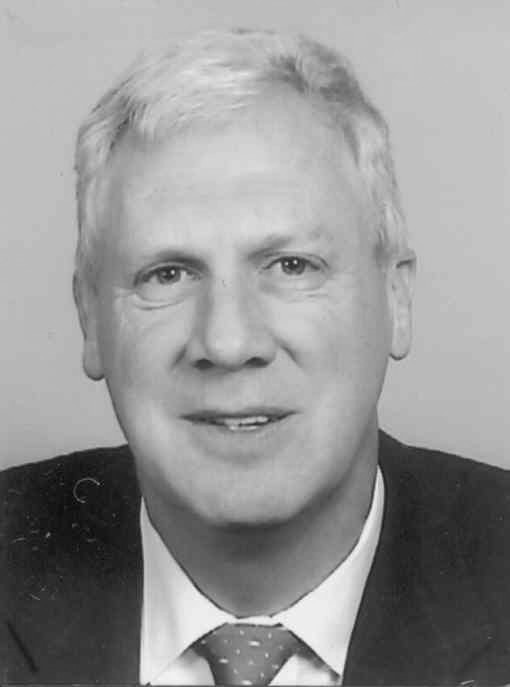}}]{Jaap van den Herik}
Jaap van den Herik is Professor of Computer Science and Law at the Leiden University, at the Faculty of Science (since 2014) and the Faculty of Law (since 1988). Previously, he was affiliated with the Tilburg University (2008-2014) and the Maastricht University (1987-2008) as full Professor of Computer Science. He is the founding Director of IKAT (Institute of Knowledge and Agent Technology) and TiCC (Tilburg center for Cognition and Communication) and was supervisor of 69 Ph.D. researchers. Van den Herik studied mathematics (with honors) at the Vrije Universiteit Amsterdam and received his Ph.D.degree at Delft University of Technology in 1983. He was active in many organizations, such as the Belgian Netherlands Association of AI, JURIX, the ICGA, and the consortium BIGGRID. He is ECCAI fellow since 2003.
Van den Herik is member of the  TWINS (the research council for sciences  of the KNAW) and a member of the Royal Holland Society of Sciences and Humanities. In 2014, he received the HUMIES Award at GECCO2014 together with David (Eli) Omid Tabibi, Nathan Netanyahu, and Moshe Koppel for their contribution: Genetic Algorithms for Evolving Computer Chess Programs.
His research interests are: Computer Games, Search Algorithms, Genetic Algorithms, Serious Games, Adaptive Agents, Neural Networks, Information Retrieval, Big Data, e-Humanities, Intelligent Crowd Sourcing, and Intelligent Systems for Law Applications.
\end{IEEEbiography}


\end{document}